\documentclass[10pt,journal,compsoc]{IEEEtran}
\usepackage{graphicx}
\usepackage{epstopdf}
\usepackage{amsmath}
\usepackage{amssymb}
\usepackage{myalgorithm,myalgorithmic}
\usepackage{multirow}
\usepackage{hyperref}
\usepackage{color}
\usepackage{subfigure}
\usepackage{epsfig}
\usepackage{array}

\begin{document}
%
\title{Tasks Integrated Networks: Joint Detection and Retrieval for Image Search}
%
\author{Lei Zhang,
        Zhenwei He,
        Yi Yang,
        Liang Wang,
        Xinbo Gao%

\IEEEcompsocitemizethanks{\IEEEcompsocthanksitem  
L. Zhang and Z. He are with the School of Microelectronics and Communication Engineering, Chongqing University, Chongqing 400044, China.\protect\\
\ (E-mail: leizhang@cqu.edu.cn, hzw@cqu.edu.cn).
 \IEEEcompsocthanksitem  Y. Yang is with the Center for Artificial Intelligence, University of Technology Sydney, Australia.
 \ (E-mail: yi.yang@uts.edu.au).
  \IEEEcompsocthanksitem  L. Wang is with the National Laboratory of Pattern Recognition, Institute of Automation, Chinese Academy of Sciences, Beijing 100190, China.
 \ (E-mail: wangliang@nlpr.ia.ac.cn).
   \IEEEcompsocthanksitem  X. Gao is with the Chongqing Key Laboratory of Image Cognition, Chongqing University of Posts and Telecommunications, Chongqing 400065.
 \ (E-mail: gaoxb@cqupt.edu.cn).

}}

%
%

\markboth{IEEE TRANSACTIONS ON PATTERN ANALYSIS AND MACHINE INTELLIGENCE, 2019}%
{Shell \MakeLowercase{\textit{et al.}}: Bare Demo of IEEEtran.cls for IEEE Journals}
%




\IEEEtitleabstractindextext{%
\begin{abstract}
The traditional object (person) retrieval (re-identification) task aims to learn a discriminative feature representation with intra-similarity and inter-dissimilarity, which supposes that the objects in an image are manually or automatically pre-cropped exactly. However, in many real-world searching scenarios (e.g., video surveillance), the objects (e.g., persons, vehicles, etc.) are seldom accurately detected or annotated. Therefore, object-level retrieval becomes intractable without bounding-box annotation, which leads to a new but challenging topic, i.e. image-level search with multi-task integration of joint detection and retrieval. In this paper, to address the image search issue, we first introduce an end-to-end Integrated Net (I-Net), which has three merits: 1) A Siamese architecture and an on-line pairing strategy for similar and dissimilar objects in the given images are designed. Benefited by the Siamese structure, I-Net learns the shared feature representation, because, on which, both object detection and classification tasks are handled. 2) A novel \textbf{o}n-\textbf{l}ine \textbf{p}airing (OLP) loss is introduced with a dynamic feature dictionary, which alleviates the multi-task training stagnation problem, by automatically generating a number of negative pairs to restrict the positives. 3) A \textbf{h}ard \textbf{e}xample \textbf{p}riority (HEP) based softmax loss is proposed to improve the robustness of classification task by selecting hard categories. The shared feature representation of I-Net may restrict the task-specific flexibility and learning capability between detection and retrieval tasks. Therefore, with the philosophy of \textbf{d}ivide and \textbf{c}onquer, we further propose an improved I-Net, called DC-I-Net, which makes two new contributions: 1) two modules are tailored to handle different tasks separately in the integrated framework, such that the task specification is guaranteed. 2) A class-center guided HEP loss (C$^2$HEP) by exploiting the stored class centers is proposed, such that the intra-similarity and inter-dissimilarity can be captured for ultimate retrieval. Extensive experiments on famous image-level search oriented benchmark datasets, such as CUHK-SYSU dataset and PRW dataset for person search and the large-scale WebTattoo dataset for tattoo search, demonstrate that the proposed DC-I-Net outperforms the state-of-the-art tasks-integrated and tasks-separated image search models.
\end{abstract}
%
}
\maketitle

\IEEEdisplaynontitleabstractindextext
\IEEEpeerreviewmaketitle

\section{Introduction}
%
%
%
%
\IEEEPARstart{S}{earching} images containing some interested object from a large gallery image set is a new but challenging research area. For example, in real-world video surveillance, many tasks such as criminals search~\cite{Wang2013Intelligent} and multi-camera tracking~\cite{Song2010Tracking} are closely related to our life. These tasks need to search the image containing a target person from the videos of different scenes and backgrounds where the persons in the videos are not cropped or annotated. Generally, for person search tasks, the machine should first know where are the persons in the image (pedestrian detection) and then guess who is the right person (person re-identification). Therefore, image search problem is closely related to two independent computer vision tasks, such as object detection (positioning) and object retrieval (matching). The detection model aims to locate the interested objects in the images, while the purpose of the retrieval model model is to match the query objects and gallery objects where the two images may come from different distributions. For example, in pedestrian detection and person re-identification problems, camera views, poses, occlusions, illuminations, backgrounds and resolutions may easily cause intra-class dissimilarity and inter-class similarity. Therefore, both detection and retrieval are challenging problems in computer vision and have attracted lots of attention in recent years~\cite{Chen2016Deep},~\cite{Cao2017Solving},~\cite{Song2017Collaborative},~\cite{Ouyang2013Modeling},~\cite{shen2018weakly},~\cite{he2017mask}.

\begin{figure}[t]
\begin{center}
   \subfigure[Person Re-identification]{
   \label{Figure 1.a}
   \includegraphics[width=1.0\linewidth]{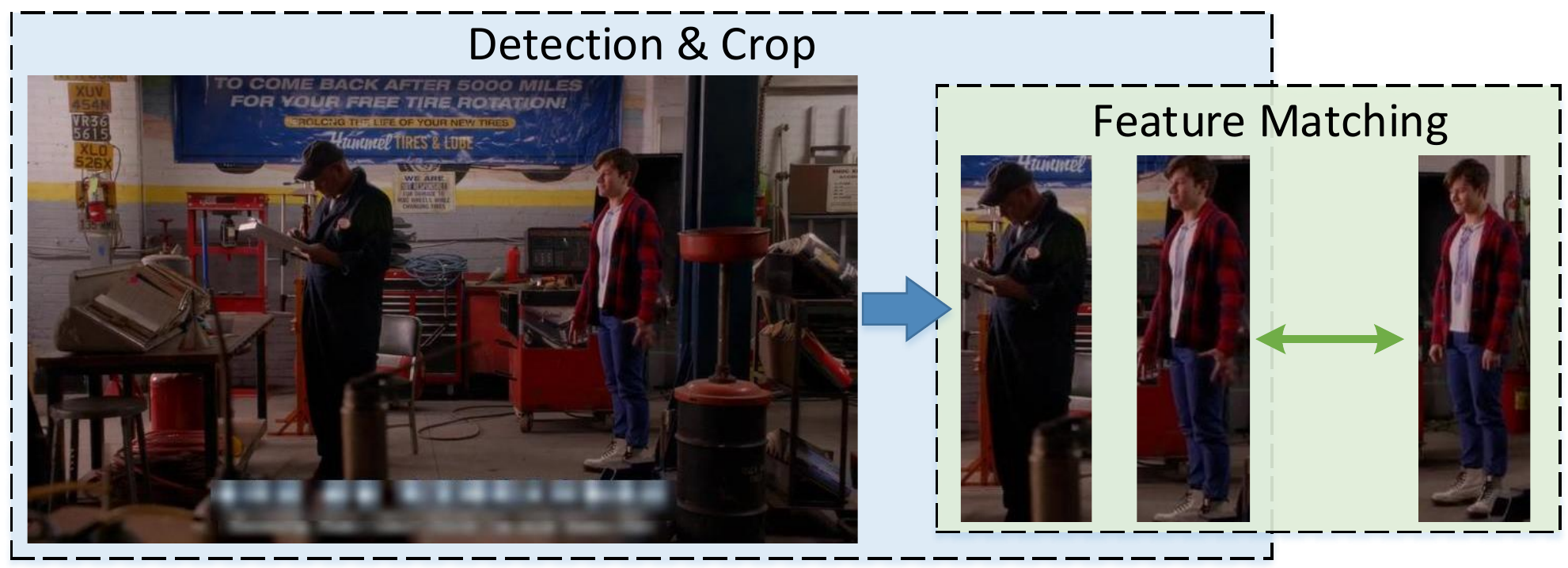}
   }
   \hspace{1in}
   \subfigure[Person Search (our work)]{
   \label{Figure 1.b}
   \includegraphics[width=1.0\linewidth]{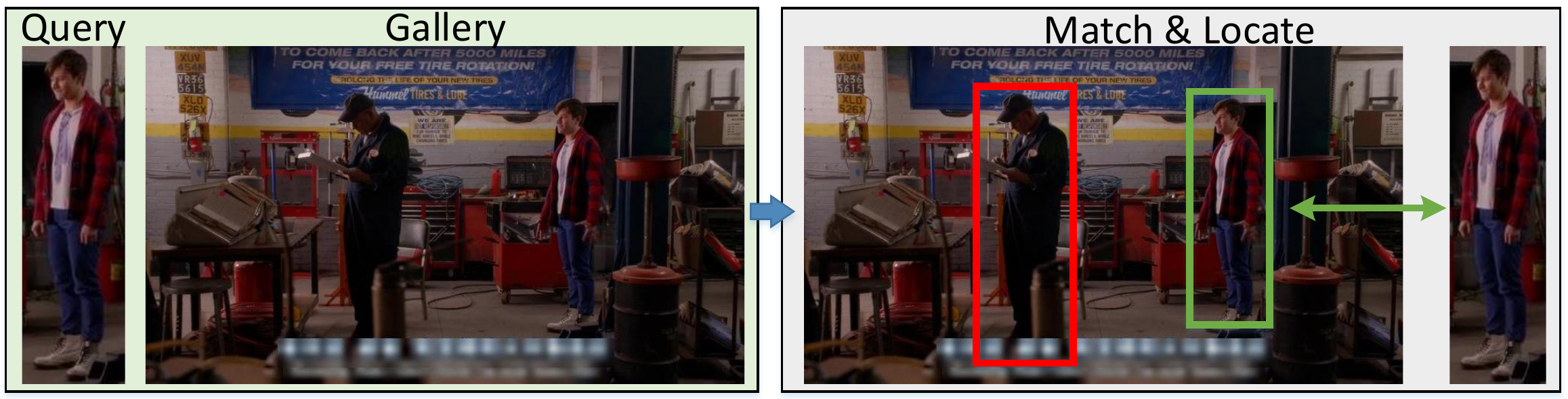}
   }
\end{center}
   \caption{Comparison of person re-identification and person search. In (a), the traditional person re-identification task needs to first detect and crop all person regions for feature matching. The person search task in (b) means the joint detection and re-identification for localization.}
\label{fig:compare}
\end{figure}

Person search is the first attempt for image search issue. Before that, although numerous endeavor on person detection and re-identification has been made, most of them handle these two problems independently. That is, the traditional methods divide the person search task into two separated subtasks. First, a pedestrian detector is implemented to predict the bounding boxes of persons from images, and then the Rectangular regions of detected persons are cropped based on the coordinates of the predicted bounding boxes. Second, the feature representations of the detected person regions of interest are computed for person re-identification. \textcolor{black}{In general person re-identification (Re-ID) task, the pedestrian images are manually annotated and cropped for discriminative feature representation network training~\cite{Chen2016Deep},~\cite{Li2012Human},~\cite{Li2013Locally}, which gives rise to two considerations. On one hand, in real-world video surveillance task, most of the detectors inevitably have the false alarms and misalignments, which, to some extent, may cause a significant performance drop of re-identification accuracy. On the other hand, the two independent subtasks seem to be less user-friendly for ultimate Re-ID in real applications. Therefore, in order to reduce the gap between the traditional person Re-ID and real-world application, we propose to deal with a person search task by jointly detecting and matching the target person when given only two images, which is more user-friendly. Person search task focuses on the co-learning of detection and person re-identification, which means the two tasks can adapt to each other in an integrated framework to outperform single-task models.}
The difference between traditional person re-identification and person search is clearly presented in Fig.~\ref{fig:compare}.

Specifically, we propose an \textbf{I}ntegrated \textbf{N}et (I-Net) which learns the detector and re-identifier in a deep unified framework end-to-end for image search task. Three important aspects should be taken into consideration. (1) a detector should be tailored for locating the objects given an image. In I-Net, the Faster-RCNN~\cite{Ren2017Faster} based two-stage detection mechanism is considered. (2) a feature representation should be learned for object retrieval. \textcolor{black}{However, the traditional metric learning loss such as triplet loss~\cite{Schroff2015FaceNet} or triplet-wise architecture are hard to be directly implemented on the end-to-end detection and retrieval integration training structure due to the lack of objects (e.g., persons) with different identities in each iteration caused by the inherent few input images of detection task. To this end, a softmax guided \textbf{o}n\textbf{l}ine \textbf{p}airing (OLP) metric loss and \textbf{h}ard \textbf{e}xample \textbf{p}riority (HEP) classification loss are proposed for intra-similarity and inter-dissimilarity learning of feature representation in a pair-wise Siamese architecture}. The joint learning of the softmax guided metric loss (OLP) and classification (HEP) loss helps to learn more discriminative representations that benefit to the person search task. (3) the detector and re-identifier should be jointly trained, because the co-learning of two modules can promote the adaptation capability of each other, the retrieval performance and simultaneously the user-friendly property in real-world application.

\begin{figure*}[t]
\begin{center}
   \includegraphics[width=1.0\linewidth]{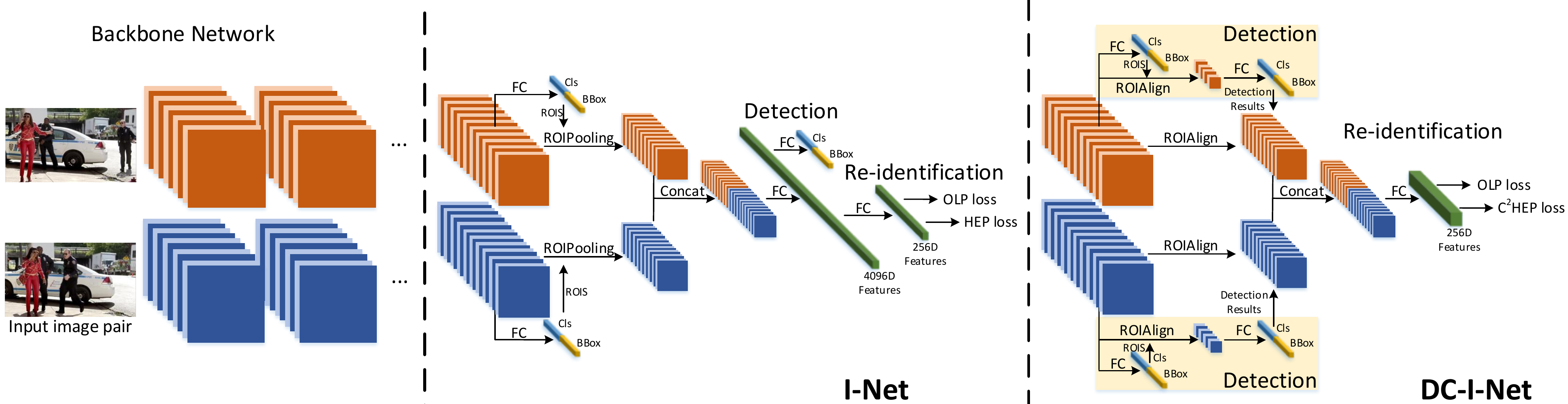}
\end{center}
   \caption{The flowchart of the proposed models. The Siamese network structure of the proposed I-Net (middle) and DC-I-Net (right) are presented, for both of which, the same backbone network (left) with shared weights is used for feature maps extraction. A pair of images containing the same objects (e.g., persons) are fed into the model for training. For I-Net, two Region Proposal Networks (i.e., Pedestrian Proposal Networks in this case) are implemented for getting the object proposals in each image, and the proposal features generated from the ROI-pooling layers are concatenated and fed into the fully-connected layers for detection and retrieval (i.e., re-identification in this case). For the DC-I-Net, the proposals of each stream are fed into the fully-connected layers for refined detection results. After that, the object features generated by the ROI-Align layers based on the refined detection results are concatenated and fed into another full-connected layer for re-identification. The essential differences between I-Net and DC-I-Net lie in two aspects. 1) For the former, detection and re-identification are treated separately in different layers. 2) For the latter, the two-stage refined objects are used for re-identification rather than the one-stage proposals.}
\label{fig:netstructure}
\end{figure*}

The structure of our I-Net is based on the Siamese network, as is shown in Fig.~\ref{fig:netstructure}. The two-stream network structure gives rise to three advantages:

1) With a Siamese structure, a pair of images can be fed into the model, such that effective feature representation with intra-similarity and inter-dissimilarity for object retrieval can be trained via joint metric and classification loss functions. More importantly, the number of input images can be enlarged, which is very helpful for better learning of the detection module (i.e., Faster-RCNN~\cite{Ren2017Faster}). Otherwise, the retrieval module will be difficult to be trained without successful detection. Additionally, with more input images, more positive pairs can be constructed to improve the metric learning of the retrieval module with better features.

2) The OLP loss with a dynamic and on-line feature dictionary is proposed, in which the stored features can help the loss function to generate negative pairs to restrict the positive pairs. Comparing to the famous triplet loss~\cite{Schroff2015FaceNet} for similarity metric learning, which is widely used in the traditional person re-identification tasks~\cite{Cheng2016Person},~\cite{Liu2016Multi},~\cite{Ding2015Deep},~\cite{song2018mask}, our loss can well remit the stagnation problem that is easily encountered in training the triplet loss. The stagnation is due to that for the end-to-end person search task, \textcolor{black}{the input image number of each iteration is so small that the number of identities for generating the positive and negative pairs is not enough for training.} In our OLP loss, an on-line feature dictionary is deployed to solve the scarcity of samples for training. Additionally, the OLP is designed with a cross-entropy formulation for similarity metric learning by computing the confidence probability of similarity. The stagnation problem can be well remitted.

3) The HEP loss is further introduced as a classification loss, which is complementary to the metric loss, i.e., OLP, because HEP loss improves the class (identity) discrimination ability of the learned network. Different from the traditional classification loss, the HEP loss selects and focuses on the hard categories via a hard example priority strategy. With the integration of the newly designed OLP and HEP for retrieval and the standard detector losses (i.e., cls. vs. reg.) in Faster-RCNN~\cite{Ren2017Faster} for detection, the proposed I-Net can be trained end-to-end for both detection and retrieval tasks, and user-friendly image search is achieved.

\textit{New Challenges and Improved Solutions of I-Net}. As shown in Fig.~\ref{fig:netstructure}, \textcolor{black}{the shared feature representation of the FC-layer is learned between detection and retrieval in I-Net, which is a little defective due to that the features for detection are focused on the discrimination between foreground and background, while the features for retrieval are focused on the discrimination among all the foregrounds. Therefore, utilizing the shared feature representation from the same layer for both detection and retrieval tasks may deteriorate the performance of image search}. Also, in I-Net, the object proposals rather than precise objects are used for retrieval, which may also degrade the performance. With the above new challenges, in this paper, we revisit the image search framework based on I-Net, and further propose new solutions by extending our I-Net~\cite{he2018end} with reconsideration of the feature shared network structure problem for joint multi-task co-training. Specifically, with the philosophy of divide and conquer, in the I-Net, we naturally deploy different layered features for each task rather than the shared representation. Specifically, an improved \textbf{d}ivide and \textbf{c}onquer I-Net with a new network structure and improved loss functions, called DC-I-Net, is proposed, as is shown in Fig.~\ref{fig:netstructure}. The main difference of DC-I-Net from the I-Net lies in that each task is performed on different feature layer, such that the more precise objects after the two-stage object detection are utilized for retrieval. Additionally, in DC-I-Net, we improve the training efficacy of HEP loss by exploiting the class centers of the priority classes, and propose a class center guided HEP, called C$^2$HEP loss.

This paper is a substantial extension of our I-Net, where we have made the following new contributions in network architecture and feature loss function:

\begin{itemize}
\item Consider the specificity of each task, with the philosophy of divide and conquer, an improved DC-I-Net with new network structure is proposed. The improved structure treats the tasks differently by specially deploying different layered features rather than the shared feature representation. Also, the features for retrieval are from precisely positioned objects instead of coarse proposals, such that the flexibility and plasticity in the co-learning of each task can be improved over the I-Net.

\item An improved HEP loss, called C$^2$HEP, is proposed for promoting the training efficacy of traditional softmax based cross-entropy loss. The C$^2$HEP loss is formulated by using the progressively updated class centers of each class computed with the timely updated input features.

\item Extensive experiments on benchmark datasets, such as CUHK-SYSU dataset and PRW dataset for person search and the large-scale WebTattoo dataset for tattoo search, demonstrate that the proposed DC-I-Net achieves another new record based on I-Net and outperforms many other state-of-the-art image search models of task-separately and jointly trained.
\end{itemize}

The remainder of this paper is organized as follows. Section~\ref{relatedwork} presents the related work. Section~\ref{inet} presents the proposed I-Net framework. The new materials of the proposed DC-I-Net is presented in Section~\ref{dcinet}. The experiments and results are presented in Section~\ref{experiment}. The model analysis and discussions of the proposed models are presented in Section~\ref{modelanalysis}. Finally, Section~\ref{conclusion} concludes this paper.
\section{Related Work}
\label{relatedwork}
This paper mainly addresses an image search issue together with person search and tattoo search. Consider that a number of \textit{person} related research work has been widely studied in computer vision in recent years, we therefore briefly revisit the closely-related pedestrian detection, person re-identification and some existing person search works.

\subsection{Pedestrian Detection}

Pedestrian detection is an important branch in object detection. In the early years, traditional object detection methods were designed based on the hand-crafted features and AdaBoost classifiers, such as ACF~\cite{Dollar2014Fast}, LDCF~\cite{Nam2014Local}, Checkerboards~\cite{Zhang2015Filtered} and Integral Channels Features (ICF)~\cite{Doll2009Integral}. In 2005, Dalal and Triggs~\cite{Dalal2005HOG} proposed the Histograms of oriented gradients (HOG) with support vector machine (SVM) classifier which first opened the research of human detection. These traditional methods dominate the field of detection for many years due to their robustness and effectiveness. Motivated by the great success of the convolutional neural networks (CNN), many deep learning based pedestrian detection methods have been developed in recent years. Tian \emph{et al.}~\cite{Tian2014Pedestrian} jointly optimized pedestrian detection with semantic tasks, including pedestrian attributes and scene attributes. Song \emph{et al.}~\cite{Song2017Collaborative} combined multiple deep networks with one fully-connected layer to improve the detection accuracy. In~\cite{Zhang2016Is}, CNN features extracted by a region proposal network (RPN)~\cite{Ren2017Faster} are fed into the random forest for pedestrian detection. As a famous two-stage detector, Faster-RCNN~\cite{Ren2017Faster} generates proposals in the first stage and refines the object for more accurate detection in the second stage, which is trained in an end-to-end manner and achieves state-of-the-art detection performance. In this paper, Faster-RCNN~\cite{Ren2017Faster} is taken into account as the object detector in the integrated image search network.

\subsection{Person Re-identification}

Person re-identification (Re-ID) aims to match a query person (probe) from a set of person candidates (gallery), where the probe and gallery are captured from non-overlapped camera views, which have attracted lots of attention in recent years~\cite{xu2018attention},~\cite{zheng2018pedestrian},~\cite{Zhong2016Rerank},~\cite{Xiao2016DGD},~\cite{zheng2017unlabeled},~\cite{yu2017cross},~\cite{zheng2015partial},~\cite{ma2013domain},~\cite{Deng2017DARE},~\cite{Liu2019CVFL}. Person Re-ID is still an open yet challenging issue for real-world surveillance application, due to the diverse variations of human poses, camera views, backgrounds, illumination, occlusions and resolutions. Earlier person re-identification milestones focus on the feature representation~\cite{Ahmed2015An},~\cite{Zheng2016A},~\cite{Chen2017A},~\cite{Wang2007Shape},~\cite{Gray2008Viewpoint} and similarity metric learning~\cite{Liao2015Person},~\cite{Roth2012Large},~\cite{Li2015Multi}. Chen \emph{et al.}~\cite{Chen2017A}, decomposed the person re-identification task into the classification and ranking subtasks, and jointly optimized them during the training phase. Cheng \emph{et al.}~\cite{Cheng2016Person} presented a novel multi-channel parts-based model with triplet loss to learn discriminative feature representation. Some of the person re-identification methods in recent years were designed based on mask learning. Xu \emph{et al.}~\cite{xu2018attention} masked out the background of the person and generated a pose-guided score. Song \emph{et al.}~\cite{song2018mask} introduced a three-stream network with triplet loss. Recently, Sun~\emph{et al.}~\cite{Sun2019PCB} proposed a part-based convolutional baseline (PCB) which fully exploited the local part-level information for feature discrimination. Additionally, in order to remit the stagnation problem of the triplet loss, Chen \emph{et al.}~\cite{Chen2017Beyond} enlarged the three-steam network to quadruplet network such that one more negative pair can be obtained to restrict the positive pairs. In this work, we propose an OLP loss by generating a number of negative samples to restrict the positive pair, which can effectively alleviate the stagnation problem and make end-to-end training much easier.

\subsection{Person Search}

\textcolor{black}{Person search, as one typical task of image search, have very recently attracted people's attention, which can be divided into two branches: individually trained models and jointly trained models. For the former, the ID-discriminative Embedding (IDE) and Confidence Weighted Similarity (CWS) were firstly proposed by Zheng \emph{et al.}~\cite{Zheng2016Person} for person search, in which the detector and re-identifier were trained individually. Recently, MSM~\cite{lan2018person}, MGTS~\cite{Di2018Person} and Local Refinement~\cite{han2019re}, also abandon the end-to-end network structure and deploy two distinct backbones for detection and person re-identification, respectively. After detection, these models convert the person search into a person re-identification (ReID) task. Although these models can work well in person search benchmarks, the intrinsic relationship between detection and re-identification is neglected. Additionally, the individually trained models lose the advantages of low computational cast and user-friendly property of person search. Instinctively, the detector and re-identifier can correspond and adapt to each other during joint training phase.}

For the latter, NPSM~\cite{Liu2017Neural} introduced a LSTM based end-to-end person search method which automatically reduces the region containing the target person from a given image. \textcolor{black}{Yan \emph{et al.}~\cite{yan2019learning} firstly introduced the GCN in person search for exploring the relation between instances in an image based on the context information and achieved SOTA performance}. Xiao \emph{et al.}~\cite{Xiao_2017_CVPR} jointly trained the detection and person re-identification parts during the training phase, in which a classical OIM loss function is introduced for person re-identification. However, the OIM loss only regards the feature learning of Re-ID as a classification problem, which may not well capture the intra-similarity and inter-dissimilarity in feature representation. In contrast, our proposed On-line Pairing (OLP) loss with a Siamese architecture can learn effective similarity metric for discriminative representation~\cite{he2018end}. The newly designed multi-task network in this paper allows it to integrate the detection loss, metric loss and classification loss to train simultaneously for more accurate and user-friendly person search.

\section{The Proposed Integrated Net (I-Net)}
\label{inet}
\begin{figure*}[t]
\begin{center}
   \includegraphics[width=1\linewidth]{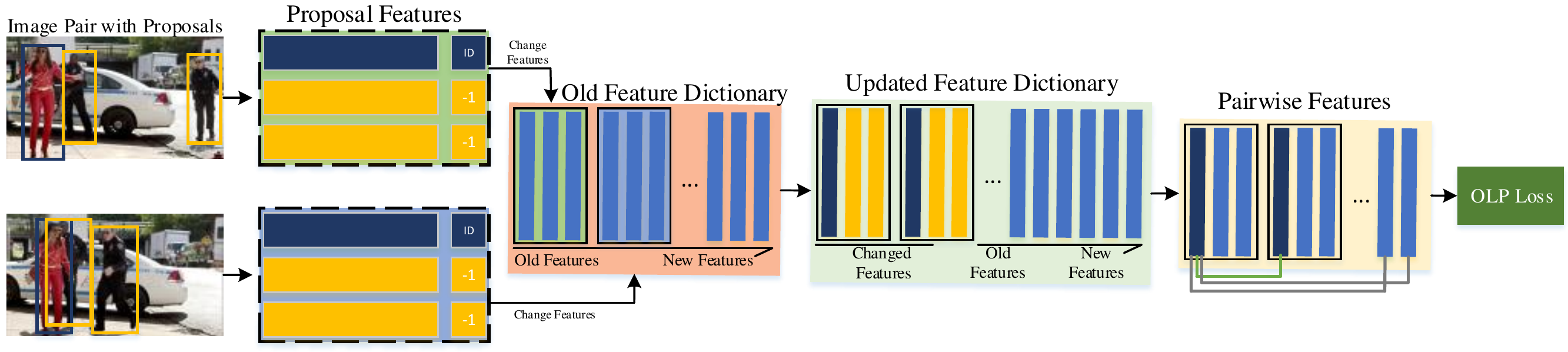}
\end{center}
   \caption{The flow-chart for computing the OLP loss. The detected proposals include two types of pedestrian, i.e., person with identity (p-w-id) in dark blue box and person without identity (p-w/o-id) in yellow box. The yellow box is labeled as -1. These proposal features are stored in the feature dictionary, which is used to construct the pair-wise features including positive pairs (green lines) and negative pairs (gray lines).}
\label{fig:OLPloss}
\end{figure*}

In this section, we will introduce our Integrated Net (I-Net) in details. The I-Net is proposed to jointly handle the detection and person re-identification into an end-to-end framework for user-friendly image search task. The Siamese architecture of I-Net is shown in Fig.~\ref{fig:netstructure} (middle). For each iteration, a pair of images containing objects (e.g., persons) of the same identity are fed into the Siamese I-Net. The backbone network is used for preliminary feature representation. After that, two region (pedestrian) proposal networks (RPN) are implemented to get the person proposals in each image, respectively. These proposals are then fed into ROI-pooling layers and the output feature maps are formulated, which is followed by two fully-connected layers for detection task and an extra 256-D $\mathcal{L}_2$-normalized feature layer for retrieval (i.e., re-identification) task. Two loss functions, i.e., OLP and HEP losses, are proposed for learning useful features with respect to re-identification.

\subsection{Novel Design of Network Structure}

\textit{Backbone}. Our I-Net is specially designed based on the Siamese structure. The backbone of I-Net is based on the VGG16~\cite{Simonyan2014Very}, which has five stacks of convolutional layers, including 2, 2, 3, 3, and 3 convolutional layers for each stack. 4 max-pooling layers are followed for the first four stacks. On the top of $conv5\_3$ layer, we generate 512 channelled feature maps for predicting the pedestrian proposals. A $512\times3\times3$ convolutional layer is first added to get the features for computing the pedestrian proposals.

\textit{Object Proposals}. Similar to the faster RCNN~\cite{Ren2017Faster}, we also associate nine anchors at each feature map. Then, a softmax classifier (cls.) supervised by cross-entropy loss is used to predict whether the anchor is a pedestrian or not, and a Smooth-$\mathcal{L}_1$ Loss (reg.) is used for bounding box regression. Finally, 128 proposals from each image after the non-maximum suppression (NMS) are obtained, which is generally recognized as the $1^{st}$ stage detection. Note that, the two branches of the Siamese network are weights shared in our model, and the two RPNs simultaneously generate the object (pedestrian) proposals from each two given input images for subsequent Re-ID task.

\textit{Joint Detector and Re-identifier}. With the generated proposals from both two RPNs, the ROI pooling layer~\cite{Girshick2015Fast} is integrated into our I-Net to pool the features from the convolution layer. The pooled features from both two branches are then fed into the two fully-connected layers with 4096 neurons. In order to remove the false positives from the pedestrian proposals, a two-class (binary) softmax classifier (cls.) for differentiating person from others (non-persons) is trained by cross-entropy loss. Note that, for general image search task, a multi-class softmax classifier will be trained for general object detection. The Soomth-$\mathcal{L}_1$ loss is used to refine the locations of bounding boxes. This is generally recognized as the $2^{nd}$ stage detection similar to Faster-RCNN. Then a pair of 256-D $\mathcal{L}_2$-normalized features for each pair-wise images generated by an extra fully-connected layer are fed into the on-line pairing (OLP) loss and hard example priority (HEP) loss for training the re-identification module, and the details of OLP and HEP will be presented in the following sections. With these two-stage detection losses (cls. vs. reg.) and re-identification losses (OLP vs. HEP), the proposed I-Net can be jointly trained for simultaneous person detection and re-identification in an end-to-end Siamese network architecture, as is shown in Fig.~\ref{fig:netstructure}.

\textit{Overall Merits}. There are two clear merits of the Siamese structure. On one hand, image search is actually an image matching problem between a query image (probe) and each gallery image, which means that the two-stream I-Net can appropriately match two given images, by training the metric loss and classification loss simultaneously. On the other hand, with the pair-wise images as inputs, the number of samples is increased, which is beneficial to the training efficacy and robustness of the detector and re-identifier.

\subsection{On-line Pairing Loss (OLP)}
\label{olp}

The OLP loss is proposed for Re-ID task by learning discriminative metric, with the consideration of two aspects:
\begin{itemize}
\item First, the features for person re-identification should be restricted by the loss function during the training phase, such that the features of the same identity should have smaller distance (intra-similarity) and the features of different identity should have larger distance (inter-dissimilarity).
\item \textcolor{black}{Second, due to the insufficient input image numbers and lock of objects in each image}, the stagnation problem of the traditional metric loss (e.g., triplet loss) easily happens because of many easy pairs but few identities (few-shot), which seriously prevents the model from being effectively trained.
\end{itemize}

With the above considerations, in the model we institutively deploy a dynamic feature dictionary to store the proposal features, such that more negative pairs can be generated together with the positive pairs drawn from more identities. As a result, an effective metric with these positive and negative pairs in a Siamese structure can be learned. The reason is that the condition of the loss function is much harder to be satisfied because of many more pairs and identities shot, and the stagnation problem in training can then be remitted.

Specifically, for each iteration, the $1^{st}$ stage detector provides 128 bounding boxes (proposals) in each image. In the person search datasets, without considering the proposals of backgrounds, there are two types of pedestrian bounding boxes, i.e., persons with identity information (p-w-id) and persons without identity information (p-w/o-id).
The detailed flow chart of the proposed OLP loss is described in Fig.~\ref{fig:OLPloss}, in which the p-w-id and p-w/o-id are represented by \textcolor{black}{dark blue} and yellow bounding boxes, respectively. We observe that an online feature dictionary of fixed size is deployed to store the generated features together with their identity label (i.e., -1 for the p-w/o-id). In this work, the number of features stored in the dictionary (i.e., dictionary size) is set as 40 times the number of bounding boxes generated by the detector. When the dictionary is filled with features during training phase, the oldest features in the dictionary will be replaced with new ones.

\textit{Formulation}. In order to minimize the discrepancy between the features of the same identity and simultaneously maximize the discrepancy between the features of different identity, we have the following notations for formulation of the OLP loss. Suppose the group of proposals for loss computation to be $(\textbf{p}_{1}, \textbf{p}_{2}, \textbf{n}_{1}, \textbf{n}_{2} ..., \textbf{n}_{K})$, where $(\textbf{p}_{1}, \textbf{p}_{2})$ stand for bounding boxes from the same identity generated by the model in forward propagation, $(\textbf{n}_{1}, \textbf{n}_{2} ... , \textbf{n}_{K})$ are the features stored in the dictionary labeled as negative samples, and $K$ means the number of negative samples for the $i^{th}$ subgroup. For each pairing group, we tend to formulate two symmetrical subgroups by taking $\textbf{p}_{1}$ and $\textbf{p}_{2}$ as anchor, alternatively. For example, when $\textbf{p}_{1}$ is regarded as anchor, then $(\textbf{p}_{1}, \textbf{p}_{2})$ denotes the positive pair, while $(\textbf{p}_{1}, \textbf{n}_{1})$, $(\textbf{p}_{1}, \textbf{n}_{2})$, ..., $(\textbf{p}_{1}, \textbf{n}_{K})$ represent negative pairs. Alternatively, when $\textbf{p}_{2}$ is regarded as anchor, then $(\textbf{p}_{2}, \textbf{p}_{1})$ denotes the positive pair, while $(\textbf{p}_{2}, \textbf{n}_{1})$, $(\textbf{p}_{2}, \textbf{n}_{2})$, ..., $(\textbf{p}_{2}, \textbf{n}_{K})$ represent negative pairs. Obviously, the OLP loss function generates more negative pairs to restrict the positive pair, which is able to remit the stagnation problem.

Suppose that we get $m$ subgroups in one iteration, and $\textbf{x}_{a}^{i}$, $\textbf{x}_{p}^{i}$, $(\textbf{x}_{n_{1}}^{i}, \textbf{x}_{n_{2}}^{i}, ..., \textbf{x}_{n_{K}}^{i})$ stand for the anchor, positive and negative features of $i^{th}$ subgroup, respectively. Then, the proposed OLP loss function is represented as follows.
\begin{equation}\label{OLP_Loss}
  \mathcal{L}_{OLP} = -\frac{1}{m}
  \sum_{i=1}^{m}
  \log_{}\frac{e^{d(\textbf{x}_{a}^{i},\textbf{x}_{p}^{i})}}
  {e^{d(\textbf{x}_{a}^{i},\textbf{x}_{p}^{i})}+\sum_{k=1}^K
  e^{d(\textbf{x}_{a}^{i},\textbf{x}_{n_{k}}^{i})}}
\end{equation}
\textcolor{black}{where $d(\cdot)$ stands for the cosine similarity between two features. It is worth noting that since our features are $\mathcal{L}_2$-normalized, the cosine similarity can be easily computed by the inner product of each two feature vectors.} From Eq. (\ref{OLP_Loss}), we observe that a softmax guided cross-entropy loss works as a metric loss, and the summation of all distances is set as the denominator so that the distance of the positive pair can compare to all negative pairs in each subgroup.

In gradient computation, we only calculate the deviation with respect to the anchor feature. Then, the deviation of the OLP loss function with respect to $\textbf{x}_{a}^{i}$ for the $i^{th}$ subgroup can be calculated as:
\begin{equation}\label{hatq}
  \frac{\partial \mathcal{L}_{OLP}}{\partial \textbf{x}_{a}^{i}} = (q^{i}-1)\textbf{x}_{p}^{i}+\sum_{k=1}^K{(\hat{q}_{k}^{i}\textbf{x}_{n_{k}}^{i})}
\end{equation}
where $q^{i}$ and $\hat{q}_{k}^{i}$ are expressed as follows.
\begin{equation}\label{q}
  q^{i} = \frac{e^{d(\textbf{x}_{a}^{i},\textbf{x}_{p}^{i})}}
  {e^{d(\textbf{x}_{a}^{i},\textbf{x}_{p}^{i})}+\sum_{k=1}^K
  e^{d(\textbf{x}_{a}^{i},\textbf{x}_{n_{k}}^{i})}}
\end{equation}
\begin{equation}\label{qi}
  \hat{q}_{k}^{i} = \frac{e^{d(\textbf{x}_{a}^{i},\textbf{x}_{n_{k}}^{i})}}
  {e^{d(\textbf{x}_{a}^{i},\textbf{x}_{p}^{i})}+\sum_{k=1}^K
  e^{d(\textbf{x}_{a}^{i},\textbf{x}_{n_{k}}^{i})}}, k=1,...,K
\end{equation}

In summary, as is shown in Fig.~\ref{fig:HEPloss} and Eq. (\ref{OLP_Loss}), the proposed OLP loss can be implemented as follows:
\begin{enumerate}
\item The features of each two input images are collected. The features $(\textbf{p}_{1}, \textbf{p}_{2})$ from the images of the same identity are constructed as positive pairs.
\item For each positive pair $(\textbf{p}_{1}, \textbf{p}_{2})$, $\textbf{p}_{1}$ and $\textbf{p}_{2}$ are set as the anchor, alternatively. The features $(\textbf{n}_{1}, \textbf{n}_{2} ... , \textbf{n}_{K})$ stored in the feature dictionary are paired with the anchor to construct negative pairs.
\item Compute the OLP loss by using Eq.~(\ref{OLP_Loss}) and its gradient by using Eq.~(\ref{hatq}) for gradient back-propagation optimization.
\item Store the input features to progressively update the feature dictionary.
\end{enumerate}

Obviously, by optimizing the proposed OLP metric loss function, \textcolor{black}{the cosine similarity $d(\textbf{x}_{A}^{i},\textbf{x}_{p}^{i})$ between the features of the same identity (intra-similarity) is maximized, while the cosine similarity $d(\textbf{x}_{A}^{i},\textbf{x}_{n}^{i})$ of different identities (inter-similarity) is minimized.} Moreover, with thousands of features progressively stored in the feature dictionary, a number of negative pairs can be generated which also effectively remit the stagnation of I-Net model training. \textcolor{black}{Note that our OLP is scenario driven and tailored for the multi-task integration network, which is essentially different from the pure metric learning loss, such as the N-pair loss \cite{Sohn2016npair} that relies on enough pair-wise training samples from enough classes. The existing metric losses cannot easily be deployed in our network for feature learning due to that the number of input training samples is too small}.

\subsection{Hard Example Priority Loss (HEP)}
\label{hep}

\textcolor{black}{The OLP loss function enables the cosine distance of positive pairs to be smaller and that of negative pairs to be larger, which does not directly regress the identity labels in the loss function.} Additionally, the traditional softmax based cross-entropy loss for classifier training does not consider the degree of difficulty of the examples in the data. With the above considerations, we further propose a hard example priority (HEP) loss function, which aims to regress the identity labels with high priority. A high priority of some identity label means that the identity is hard to be classified, and will be selected for label regression and loss computation.

\begin{figure}
\begin{center}
   \includegraphics[width=1.0\linewidth]{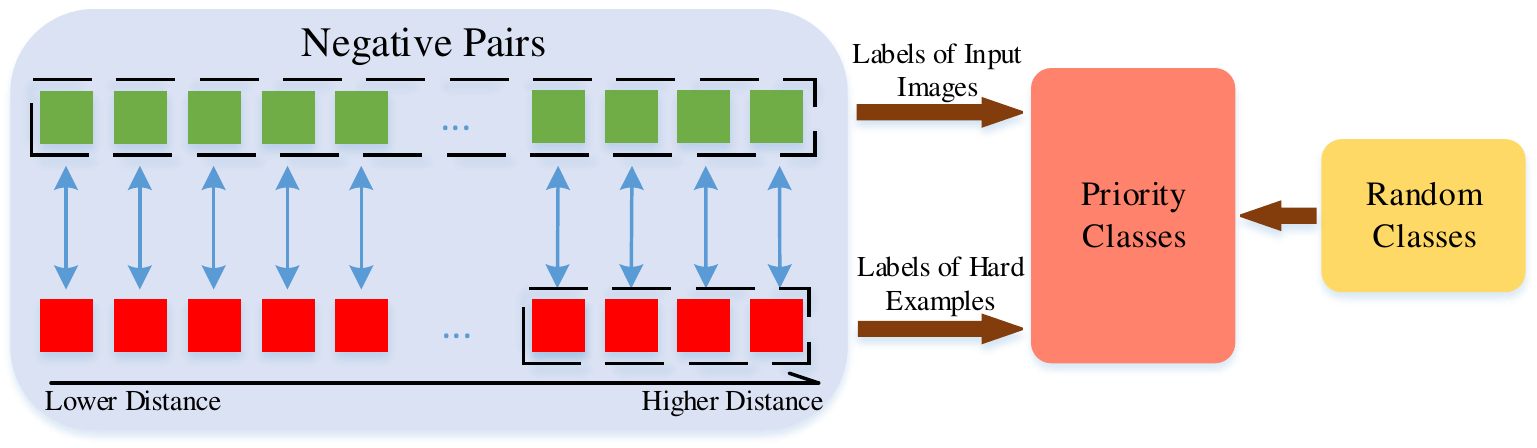}
\end{center}
   \caption{The protocol for selecting the priority classes of hard examples for computing the HEP loss. First, the person proposals (bounding boxes) with identities (i.e., ground-truth labels) are marked. Second, the negative pairs with the largest cosine distances were selected as hard examples, which are denoted as priority classes. Finally, if the pool of priority classes is not yet filled, some random classes are selected to fill the pool, which are used to compute the HEP loss.}
\label{fig:HEPloss}
\end{figure}

Suppose that there are $C$ identities. The HEP loss function aims to classify the person proposals with identity (i.e., p-w-id) into $C+1$ classes containing an extra background class. In order to calculate the HEP loss, the person proposals with high Interaction-over-Union (IOU) between the bounding boxes and the ground truth are used. Then, by computing the cosine similarity between positive pairs and negative pairs via the OLP loss as described in Section~\ref{olp}, we can determine the top $r$ maximum distances of the negative pairs. The examples with maximum distances of negative pairs (i.e., high inter-similarity) are recognized as hard examples of priority classes. In order to keep the total number of priority classes fixed, we also randomly select an uncertain number of classes from the remaining categories of non-priority classes. As a result, totally $T (T<<C+1)$ classes are selected to compute the HEP loss. In summary, suppose that the selected hard categories are stored in the pool $\mathcal{P}$, as is shown in Fig.~\ref{fig:HEPloss}, the protocol for selecting the hard categories is presented as follows:

\begin{enumerate}
\item The label indexes of each input image pairs with identities are first determined to ensure the ground-truth classes.

\item For each subgroup (described in Section~\ref{olp}), the label indexes of negative samples from the top $r$ negative pairs with the maximum distances are stored in the priority classes pool $\mathcal{P}$, which enables the priority classes of hard examples to be focused.

\item If the size of pool $\mathcal{P}$ is still smaller than the preset $T$, we randomly select several classes to fill the pool.
\end{enumerate}

Finally, with the traditional softmax based cross-entropy loss and the selected priority classes, the proposed HEP loss function is formulated as:
\begin{equation}\label{HEPLoss}
    \mathcal{L}_{HEP} = -\frac{1}{n}
  \sum_{i=1}^{n}
  \sum_{j\in \mathcal{P}}
  \textbf{1}(label = j)\log_{}\frac{e^{s_{j}^{i}}}
  {\sum_{t=1}^{T}
  e^{s_{t}^{i}}}
\end{equation}
where $s_{j}^{i}$ stands for the $i$-th proposal's score from the classifier and $j$ stands for the $j$-th class. In the loss function, only the selected categories are used for the loss computation, such that, the loss function focus on the hard category.

\subsection{Overall Loss of I-Net}
I-Net is an end-to-end model which has integrated the detection and re-identification jointly for training. Therefore, the losses are composed of two parts: detection loss ($\mathcal{L}_{Det}$) and re-identification loss ($\mathcal{L}_{OLP}$ and $\mathcal{L}_{HEP}$), which are represented as follows.
\begin{equation}\label{I_Net_Loss}
    \mathcal{L}_{I-Net} = \mathcal{L}_{Det}+\alpha \mathcal{L}_{OLP}+\beta \mathcal{L}_{HEP}\\
\end{equation}
where $\alpha$ and $\beta$ represent the trade-off parameters.

Note that the detection loss $\mathcal{L}_{Det}$ is following the traditional Faster-RCNN detector, which includes two stages and each stage includes a softmax based cross-entropy classification loss (cls.) and smooth-$\mathcal{L}_1$ based regression loss (reg.) for bounding box prediction. The training of model (\ref{I_Net_Loss}) is end-to-end for user-friendly image search task.

\begin{figure}
\begin{center}
   \includegraphics[width=1.0\linewidth]{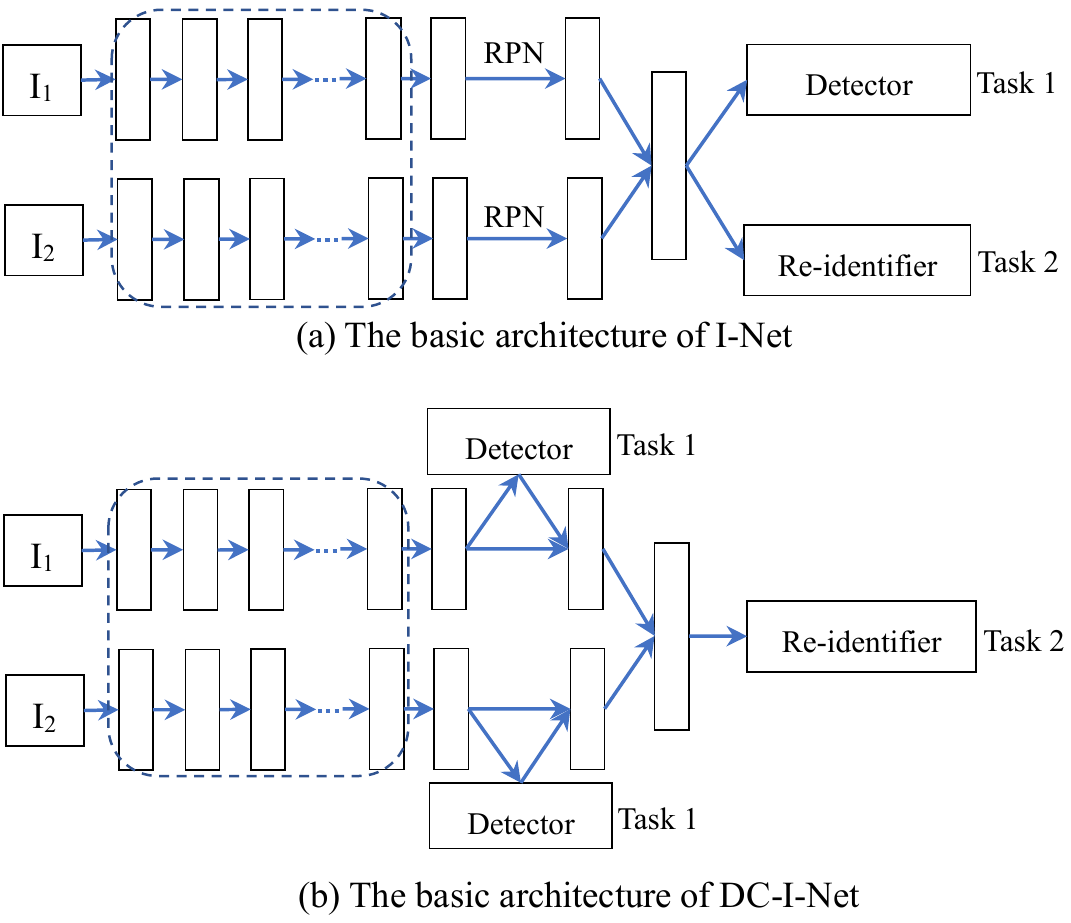}
\end{center}
   \caption{Motivation of DC-I-Net and its differences from I-Net that 1) the detector is deployed in front of the re-identifier in different layers and 2) the refined objects instead of the coarse proposals from RPN are used to train the re-identifier.}
\label{motivation_dcinet}
\end{figure}

\section{Divide and Conquer Integrated Network (DC-I-Net)}
\label{dcinet}
\textit{Motivation}. With the philosophy of divide and conquer, this section presents an improved I-Net, i.e., DC-I-Net, for dealing with several flaws of I-Net in network architecture and training loss.
The difference in architecture between the I-Net and DC-I-Net is clearly shown in Fig.~\ref{motivation_dcinet}, which, specifically, is motivated and improved from the following three important aspects:
\begin{itemize}
\item \textit{Task specialization in architecture for detection and Re-ID}. We propose a novel network structure over the I-Net, in which the features for detection and re-identification are extracted from different layers rather than the shared layer as I-Net does, by deploying the detector in front of the re-identifier. \textcolor{black}{The reason is that the features for the detection task should focus on the discrimination between the foregrounds (objects) and backgrounds, while the features for re-identification should focus on the discrimination among the foregrounds (e.g., persons of different identities)}. The novel network architecture deployed with detector and re-identifier is trained end-to-end, such that the image search task becomes more user-friendly and practical.
\item \textit{Refined object proposals for the metric loss}. In I-Net, the features from the coarse object proposals (the $1^{st}$ stage detection) are used for re-identification loss optimization, which may degrade the final retrieval performance due to the inaccurate proposals. Therefore, in DC-I-Net, the refined objects with ROI-Align (the $2^{nd}$ stage detection) are used for better training the re-identification metric loss.
\item \textit{Easy training of the classification loss}. As mentioned in Section~\ref{olp}, \textcolor{black}{due to the small input number of images during the training phase}, a very few number of identities contribute to the model training such that the model has to experience thousands of epoches for probably seeing all labeled identities. This naturally leads to the hard training of the HEP loss. In order to address this issue, we further propose a class center guided hard example priority (C$^2$HEP) loss by fully exploiting the updated input features to compute the class centers. As a result, the identity discrimination of features is much improved.
\end{itemize}

From the perspectives of network architecture and loss function, these new improvements of DC-I-Net over I-Net are presented in the following subsections in detail.

\subsection{Improved Network Architecture Over I-Net}

The backbone of DC-I-Net is the same as I-Net, as shown in Fig.~\ref{motivation_dcinet} (the dashed box). The detailed network structure of DC-I-Net is shown in Fig.~\ref{fig:netstructure} (right), which is different from the I-Net shown in Fig.~\ref{fig:netstructure} (middle) that 1) the task specialization for detection and Re-ID is well considered by using features from different layers, 2) the refined objects from the $2^{nd}$ stage detector are generated by using ROI-Align module for training the metric loss, and 3) a class-center guided hard example priority (C$^2$HEP) loss is proposed for easy training of the identity classification loss.

\textit{Detector}. In the DC-I-Net, the features for detection and person re-identification tasks are extracted from different layers. After the two-stage detection supervised by classification loss (cls.) and regression loss (reg.), the detection of accurate bounding boxes (i.e. objects) is completed.

\textit{Re-identifier}. After the two-stage detection, the coordinates of the refined bounding boxes are fed into the ROI-Align layers to compute the features of refined object proposals for person re-identification (person search) or object retrieval (image search). The pooled feature maps have a size of $7 \times 14$ for person search task, which has a similar aspect ratio to the bounding boxes of a person. The feature maps are then fed into the fully-connected layers to learn the feature vector representation for person re-identification. \textcolor{black}{Finally, 256-D $\mathcal{L}_2$-normalized features are generated for the object proposals by an extra fully-connected layer, which are then fed into the OLP loss and C$^2$HEP loss for formal training of the re-identification module}.

\subsection{Class Center Guided HEP Loss (C$^2$HEP)}

The class-center guided HEP loss (C$^2$HEP) is proposed to improve the hard training problem of HEP in I-Net, \textcolor{black}{which is an inherent problem of person search task caused by the insufficient number of identities of each iteration. Therefore, we exploit the features extracted from the input images to compute the class center of each category.} Consider that the cosine similarity between each sample and the class centers can clearly reflect the probability of the sample belonging to a class, we propose to feed the cosine similarity into the softmax function of HEP and formulate the C$^2$HEP. \textcolor{black}{For convenience, we design a class center dictionary indicated by their ground-truth label. In every iteration, the stored class center for class $j$ is updated by new features, formulated as
\begin{equation}\label{center_update}
\textbf{c}_{new}^j=\phi \cdot \textbf{c}_{old}^j + (1-\phi)\cdot \textbf{x}^{(j)}\\
\end{equation}
where $\textbf{c}_{old}^j$ is the old center of class $j$, $\textbf{x}^{(j)}$ is the input feature of class $j$, and $\phi$ ($0<\phi<1$) is a hyper-parameter, which is set as 0.5 in our implementation}. It is worth noting that the class center dictionary is different from the feature dictionary in the OLP loss. In order to fully explore the information of identities/categories, each labeled object (e.g., person) in the dataset has been assigned a class center stored in the class center dictionary.

Suppose that the feature of the $i^{th}$ object fed into the C$^2$HEP loss function is $\textbf{x}_{i}$, the class center dictionary is defined as $\mathcal{S}$, where $\mathcal{S} = \{\textbf{c}_{1}, \textbf{c}_{2}, ..., \textbf{c}_{C}\}$ and $C$ is the number of the identities (categories) annotated in the dataset. Generally, the feature more probably belongs to the class that has the smallest cosine distance between the class center and the feature. Based on the softmax function, we define the probability of $\textbf{x}_{i}$ belonging to the class $j$ is defined as:
\begin{equation}\label{proba}
p_{j} = \frac{e^{\lambda d(\textbf{x}_{i}, \textbf{c}_{j})}}
{\sum_{c\in \mathcal{P}} e^{\lambda d(\textbf{x}_{i}, \textbf{c}_{c})}}
\end{equation}
where the $\lambda$ is a hyper-parameter, \textcolor{black}{which is set as 10 in implementation} and $\mathcal{P}$ is the pool of selected priority classes, which is presented in Section~\ref{hep}. From Eq.~(\ref{proba}), we know that the highest probability $p_j$ can be achieved when a given sample $\textbf{x}_i$ belonging to the class $j$ has the smallest cosine distance to the class center $\textbf{c}_j$.

Suppose that the probability of a sample belonging to their corresponding ground truth label is represented as $\{p^{1}, p^{2},..., p^{n}\}$, where $n$ is the number of samples in one iteration. Under the assumption of independently and identically distribution of the learned features, it is rational to maximize the likelihood function: $L=\prod_{i}^{n} p^{i}$ for model training. In order to train the deep learning model, we minimise the negative log-likelihood function: $-\log L = -\log\prod_{i}^{n}(p^{i})$. With the negative log-likelihood function, the proposed C$^2$HEP loss function can be written as:
\begin{equation}\label{CHEPLoss}
    \mathcal{L}_{C^2HEP} = -\frac{1}{n}
  \sum_{i=1}^{n}
  \sum_{j \in \mathcal{P}}
  \textbf{1}(label = j)\log_{}\frac{e^{\lambda d(\textbf{x}_{i}, \textbf{c}_{j})}}
  {\sum_{l \in \mathcal{P}}
  e^{\lambda d(\textbf{x}_{i}, \textbf{c}_{l})}}
\end{equation}

From Eq.~(\ref{CHEPLoss}), we know that minimizing the loss function can effectively can constrain the cosine similarity $d(\cdot)$ between each feature and its corresponding class center to be larger. Therefore, the identity discrimination is guaranteed. \textcolor{black}{Note that the proposed scenario-driven C$^2$HEP loss is tailored for the detection and re-identification integration network, which is explicitly and implicitly different from the existing class-center based feature learning loss, such as center loss \cite{Wen2016classcenter} and prototype loss \cite{Snell2017proto}, that only relies on the available training samples. However, in person search scenario, the number of training samples in each iteration is very small (i.e. 2 samples for each iteration in our model), and the center loss and prototype loss will encounter stagnation problem. Therefore, a class-center dictionary is deployed in C$^2$HEP for effectively alleviating the training stagnation problem and improving the training efficiency. The C$^2$HEP loss can be recognized as a seamless connection to the OLP loss for feature discrimination.}

\subsection{Overall Loss of DC-I-Net}

The DC-I-Net is also an end-to-end model for user-friendly image search task. Similar to I-Net, the losses of DC-I-Net consist of the detection loss and re-identification loss. Specifically, the detection loss $\mathcal{L}_{Det}$ is following the traditional Faster-RCNN with two-stage detection, and each stage refers to the softmax based cross-entropy classification loss (cls.) and smooth-$\mathcal{L}_1$ regression loss (reg.) for bounding box prediction.
Since the two streams of the Siamese network share the same parameters for the detection, the two input images are used to train the detector simultaneously.

Besides, \textcolor{black}{the re-identification loss consists of the proposed OLP metric loss and C$^2$HEP identity classification losses for discriminative feature representation}. So the overall loss of the DC-I-Net is presented as follows:
\begin{equation}\label{TotalLoss}
    \mathcal{L}_{DC-I-Net} = \mathcal{L}_{Det} + \alpha \mathcal{L}_{OLP} + \beta \mathcal{L}_{C^2HEP}
\end{equation}
where $\alpha$ and $\beta$ stands for the trade-off parameters, \textcolor{black}{and $\mathcal{L}_{OLP}$ is given in Eq.~(\ref{OLP_Loss})}. By using the mini-batch SGD optimization, our model can be trained end-to-end for person search. In summary, four steps are involved during the training in every iteration:
\begin{enumerate}
\item \textit{Detection loss computation}. The pair-wise input images are fed into the Siamese structure for detection first, and the loss of detection for each image is computed.

\item \textit{OLP loss computation}. The detected objects (feature maps) are fed into the ROI-Align layer to get the features for re-identification task. The features are paired, and the distances of positive and negative pairs are calculated. Then, the OLP loss is computed with dynamic update of the feature dictionary.

\item \textit{C$^2$HEP loss computation}. The distances of positive and negative pairs are fed into the C$^2$HEP loss with selected priority classes. After computation of the loss, the class centers are progressively updated via the input new features.

\item \textit{Gradient computation}. Based on the computation of all losses, the model is optimized with SGD, until convergence.
\end{enumerate}

\section{Experiments}
\label{experiment}
To evaluate the effectiveness of our approaches, we conduct massive experiments on three benchmark datasets, including CUHK-SYSU dataset~\cite{Xiao2016End}, PRW dataset~\cite{Zheng2016Person} and Webtattoo dataset~\cite{han2019tattoo}, for image search tasks.
The first two datasets focus on the person image search (i.e., person search), which refers to joint object (person) detection and person re-identification tasks in our models. The third dataset serves for tattoo image search (i.e., tattoo search), which refers to joint object (tattoo) detection and image retrieval tasks in our models.
In this section, the experimental setup and experimental results for each dataset are presented.

\subsection{Experimental Setup}

\subsubsection{Implementation Details}
The proposed I-Net and DC-I-Net are implemented on Caffe~\cite{Jia2014Caffe} and py-faster-rcnn~\cite{Ren2017Faster} platform for model training and evaluation. The VGG-16~\cite{Simonyan2014Very} is used as the backbone network of our models and the pre-trained model in~\cite{Xiao2016End} is taken into account for network parameters initialization. The first two stacks of convolutional layers are frozen during the training of our models. The two branches of the Siamese network share the same parameters for both initialization and training. The RPN part of each branch generates 128 proposals for each image, and the proposals labeled as background are not useful and therefore dropped for object retrieval task. In both I-Net and DC-I-Net, the trade-off parameters $\alpha$ and $\beta$ are set as 1. The learning rate is initialized to 0.001, and drops to 0.0001 after 40k iterations. Totally, 70k iterations are set to enable convergence.

\subsubsection{CUHK-SYSU Dataset}
The CUHK-SYSU dataset~\cite{Xiao2016End} is a large dataset for person search, which contains 18184 images from the hand-held cameras and movie snapshots with large variations in viewpoint, lighting, resolution, \emph{etc}. From the annotations, there are 8432 different person identities and 96143 bounding boxes. Each labeled person has at least two images from different viewpoints. For the training/testing split, the developer of this dataset provided 11206 images of 5532 identities for training and 6978 images of 2900 identities for the test. Specifically, we follow the same experimental protocols as~\cite{Xiao2016End} for fair comparison.

\subsubsection{PRW Dataset}
The PRW dataset~\cite{Zheng2016Person} is drawn from a 10 hours of video captured by six cameras, in which five of them are $1080\times1920$ HD and the remaining one is $576\times720$ SD. Totally 11816 frames are manually annotated and results in 43110 pedestrian bounding boxes, in which 34304 pedestrians are annotated by 932 IDs. For the training/testing split, the PRW dataset provides 5134 frames of 482 labeled identities for training and 6112 frames of 450 labeled identities for testing. The task for this dataset allows the model to search a query target person (probe) from the whole testing set (gallery), which remains to be a challenging problem.

\subsubsection{Webtattoo Dataset}
The Webtattoo dataset~\cite{han2019tattoo} was presented in different viewpoints and illuminations, which consists of three parts: (i) the first part is a combination of three small-scale (less than 10K) tattoo datasets, such as Tatt-C~\cite{Ng2015Tattoo}, Flickr~\cite{Xu2016Tattoo} and DeMSI~\cite{Hrkac2016Tattoo}. (ii) The second part is a collection over 300K distracter tattoo images from the Internet. (iii) The third part is the 300 tattoo sketches drawn by volunteers. In this Webtattoo dataset, three tasks including the detection, tattoo search and sketch based tattoo search are deployed. In this paper, we focus on the joint tattoo detection and image search. Specifically, 1428 images of 400 tattoo classes are used for model training. For comparing the detection performance of different models, 755 images from 200 tattoo classes with ground-truth bounding boxes are used. For comparing the search (retrieval) performance of different models, the query set containing 200 images (one image per tattoo class) is used to search the images from a gallery set containing 355 tattoo images.

\subsection{Experiments on CUHK-SYSU Dataset}
\subsubsection{Compared Methods}
\textit{Baselines: Separated Detection and Re-ID Models}. In this section, we perform the experiments on the CUHK-SYSU dataset to investigate the effectiveness of our models. Consider that the proposed models in this dataset aim to jointly learning pedestrian detection and person re-identification, we therefore select three pedestrian detection methods and five person re-id approaches for baseline comparisons, which then result in 15 baselines for person search task. Specifically, three baseline detection methods, CCF~\cite{Yang2015Convolutional}, Faster-RCNN~\cite{Ren2017Faster} with Resnet50~\cite{He2016Deep} and ACF~\cite{Dollar2014Fast}, are used for detecting pedestrians. Besides, we also use the ground truth bounding boxes of the test set as the upper bound of the detector's performance. For the baseline re-identification methods, we evaluate several famous re-id feature representation methods including DenseSIFT-ColorHist (DSIFT)~\cite{Zhao2013Unsupervised}, Bag of Words (BoW)~\cite{Zheng2015Scalable}, Local Maximal Occurrence (LOMO)~\cite{Liao2015Person} and ID-Net(The re-identification part of OIM~\cite{Xiao_2017_CVPR}). The metric learning methods, i.e. KISSME~\cite{Roth2012Large} and XQDA~\cite{Liao2015Person} together with these feature representation are used for Re-ID. These separated detection and Re-ID methods are combined for person search, which are therefore treated as baselines in comparisons.

\textit{State-of-the Art (SOTA): Joint Detection and Re-ID Models}. To the best of our knowledge, there is only a few work on the joint training of detector and re-identifier for person search task, such as the OIM model~\cite{Xiao_2017_CVPR}, the end-to-end model (initialized model)~\cite{Xiao2016End}, NPSM~\cite{Liu2017Neural}, IAN~\cite{Xiao2017IAN}, RCAA~\cite{chang2018rcaa} and \textcolor{black}{Context Graph~\cite{yan2019learning}}. Therefore, these end-to-end person search methods are selected as the SOTA competitor of our I-Net and DC-I-Net. \textcolor{black}{Additionally, the CNN$_{v}$+MGTS~\cite{Di2018Person} which trains the detector and person re-identifier separately is also compared, because of their excellent performance}. All the experiments are following the same experimental protocols for fair comparisons and the gallery size is set as 100. \textcolor{black}{Note that the I-Net is implemented with VGG16 while the DC-I-Net is implemented with both VGG16 and Resnet50 in the experiment, because almost all the compared deep models are based on Resnet50 backbone.}

\begin{table}[t]
\begin{center}
\caption{\textcolor{black}{Comparisons of baselines, SOTA methods and our models on the CUHK-SYSU dataset}}
\label{Table:res}
\begin{tabular}{|c|m{3cm}|m{1.5cm}|m{1.5cm}|}
\hline
Detector & Re-id Method & mAP(\%) & Top-1(\%) \\
\hline
\hline
\multirow{5}*{ACF} & DSIFT~\cite{Zhao2013Unsupervised}+Euclidean & 21.7 & 25.9 \\
~ & DISFT~\cite{Zhao2013Unsupervised}+KISSME~\cite{Roth2012Large} & 32.3 & 38.1 \\
~ & BOW~\cite{Zheng2015Scalable}+KISSME~\cite{Roth2012Large} & 42.4 & 48.4 \\
~ & LOMO~\cite{Liao2015Person}+XQDA~\cite{Liao2015Person} & 55.5 & 63.1 \\
~ & IDNet~\cite{Xiao_2017_CVPR} & 56.5 & 63.0 \\
\hline
\multirow{5}*{CCF} & DSIFT~\cite{Zhao2013Unsupervised}+Euclidean & 11.3 & 11.7 \\
~ & DISFT~\cite{Zhao2013Unsupervised}+KISSME~\cite{Roth2012Large} & 13.4 & 13.9 \\
~ & BOW~\cite{Zheng2015Scalable}+KISSME~\cite{Roth2012Large} & 26.9 & 29.3 \\
~ & LOMO~\cite{Liao2015Person}+XQDA~\cite{Liao2015Person} & 41.2 & 46.4 \\
~ & IDNet~\cite{Xiao_2017_CVPR} & 50.9 & 57.1 \\
\hline
\multirow{5}*{CNN} & DSIFT~\cite{Zhao2013Unsupervised}+Euclidean & 34.5 & 39.4 \\
~ & DISFT~\cite{Zhao2013Unsupervised}+KISSME~\cite{Roth2012Large} & 47.8 & 53.6 \\
~ & BOW~\cite{Zheng2015Scalable}+KISSME~\cite{Roth2012Large} & 56.9 & 62.3 \\
~ & LOMO~\cite{Liao2015Person}+XQDA~\cite{Liao2015Person} & 68.9 & 74.1 \\
~ & IDNet~\cite{Xiao_2017_CVPR} & 68.6 & 74.8 \\
\hline
\multirow{5}*{GT} & DSIFT~\cite{Zhao2013Unsupervised}+Euclidean & 41.1 & 45.9 \\
~ & DISFT~\cite{Zhao2013Unsupervised}+KISSME~\cite{Roth2012Large} & 56.2 & 61.9 \\
~ & BOW~\cite{Zheng2015Scalable}+KISSME~\cite{Roth2012Large} & 62.5 & 67.2 \\
~ & LOMO~\cite{Liao2015Person}+XQDA~\cite{Liao2015Person} & 72.4 & 76.7 \\
~ & IDNet~\cite{Xiao_2017_CVPR} & 73.1 & 78.3 \\
\hline
\hline
\multicolumn{2}{|c|}{End-to-End(Initialized model)~\cite{Xiao2016End}} & 55.7 & 62.7 \\
\multicolumn{2}{|c|}{OIM~\cite{Xiao_2017_CVPR}} & 75.5 & 78.7 \\
\multicolumn{2}{|c|}{IAN~\cite{Xiao2017IAN}} & 76.3 & 80.1 \\
\multicolumn{2}{|c|}{NPSM~\cite{Liu2017Neural}} & 77.9 & 81.2 \\
\multicolumn{2}{|c|}{RCAA~\cite{chang2018rcaa}} & 79.3 & 81.3 \\
\multicolumn{2}{|c|}{CNN$_{v}$+MGTS~\cite{Di2018Person}} & 83.0 & 83.7 \\
\multicolumn{2}{|c|}{I-Net} & 79.5 & 81.5 \\
\multicolumn{2}{|c|}{Context Graph~\cite{yan2019learning}} & 84.1 & \textbf{86.5} \\
\hline
\multicolumn{2}{|c|}{DC-I-Net(VGG16)} & 83.7 & 85.8 \\
\multicolumn{2}{|c|}{DC-I-Net(Resnet50)} & \textbf{86.2} & \textbf{86.5} \\
\hline
\end{tabular}
\end{center}
\end{table}

\begin{figure*}[t]
\begin{center}
   \includegraphics[width=1.0\linewidth]{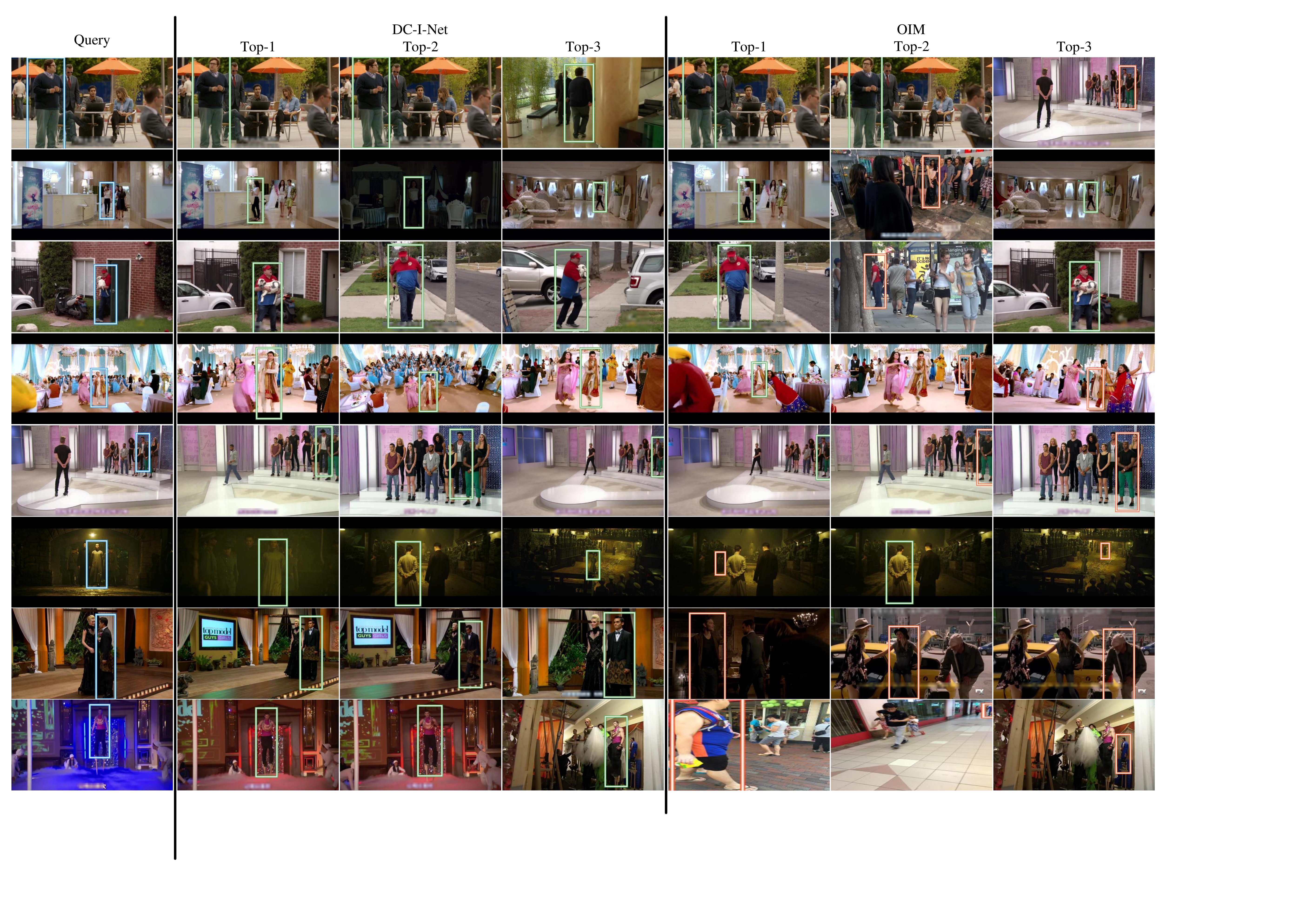}
\end{center}
   \caption{Retrieval of our DC-I-Net model (middle) and OIM (right) with the CUHK-SYSU dataset by given eight queries (left). The top 3 images with respect to the highest similarity scores are shown. The blue boxes represent the target query person (probe), the green boxes mean correct matches, and the red boxes mean incorrect matches.}
\label{vis1}
\end{figure*}

\begin{figure*}[t]
\begin{center}
   \includegraphics[width=1.0\linewidth]{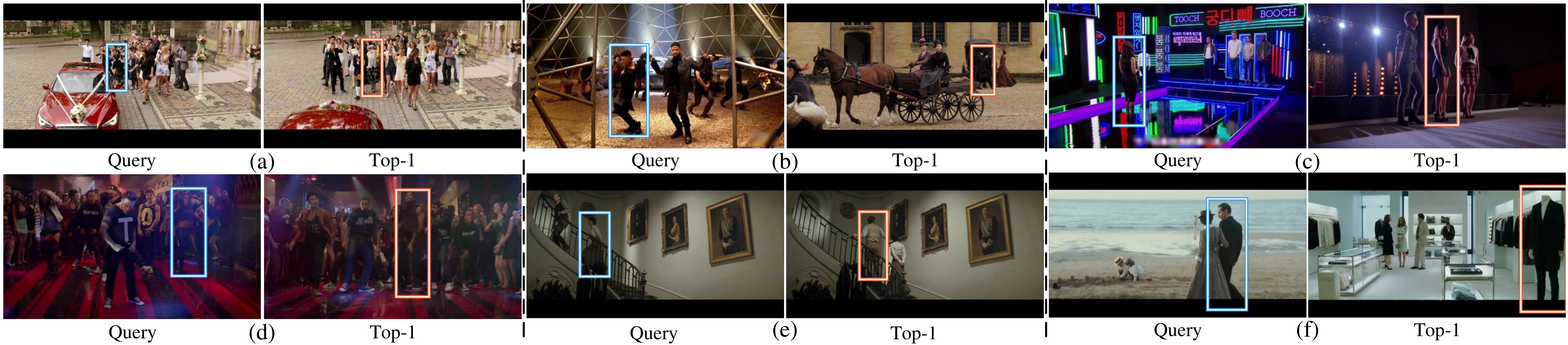}
\end{center}
   \caption{Some failures on Top-1 of our DC-I-Net model. The impacts from many real-world factors are shown, including the crowded persons (a, d), inadequate illumination (d, e), specific pose (b), false detection (f) and similar clothes (c, e), which also claim the challenges of person search.}
\label{vis2}
\end{figure*}
\subsubsection{Experimental Results}
In the experiments, the top-1 accuracy and the mAP (mean average precision) are computed for evaluating the performance of person search. Specifically, the results of person search are shown in Table~\ref{Table:res}, from which we can see that \textcolor{black}{the proposed DC-I-Net with Resnet50 achieves a top-1 accuracy of 86.5\% and mAP of 86.2\%, and outperforms all the compared methods including the SOTA context graph \cite{yan2019learning} (2.1\% in mAP) that firstly deploys GCN inside}. It is worthy noting that, with VGG16 backbone the proposed DC-I-Net outperforms the SOTA end-to-end person search model (i.e., I-Net) by 4.3\% and 4.2\% in top-1 accuracy and mAP, respectively. From Table~\ref{Table:res}, we can also observe that the SOTA end-to-end person search methods, such as OIM, NPSM, OP-I-Net, RCAA, and IAN always outperform the traditional person search methods that train the detector and person re-identifier separately rather than a joint way. It is also noteworthy that even with the ground-truth bounding boxes of pedestrians, the traditional re-identification methods have still shown inferior results compared to the end-to-end detection and Re-ID methods. This demonstrates that the joint training of both detection and re-identification modules is effective and necessary.

Additionally, the recently proposed CNN$_{v}$+MGTS~\cite{Di2018Person} is a specially designed model for person search task, which actually trains the detector and re-identifier individually. From the results, we observe that CNN$_{v}$+MGTS outperforms other compared separately or jointly trained models. However, it is still inferior to our DC-I-Net by 3.2\% and 2.8\% in top-1 accuracy and mAP, which further proves our perspective that the joint training of multiple tasks is beneficial to the between-task collaboration and improving the final performance.

\subsubsection{Evaluation Remarks}
Our model learns the feature representation of person re-identification based on both verification (OLP) and classification (C$^2$HEP) loss functions. Benefited from the Siamese structure and the feature dictionary, our model can easily generate a number of positive pairs and negative pairs to train the OLP metric loss for Re-ID task. Also, the proposed C$^2$HEP loss function can well exploit the progressively updated class centers of each category such that the input features can be trained with better identity classification capability. With these advantages, the proposed DC-I-Net model outperforms the earliest OIM~\cite{Xiao_2017_CVPR} in 2017 for person search by 8.2\% and 7.1\% in mAP and top-1 accuracy, respectively. Additionally, our DC-I-Net also outperforms the very recent recent RCAA~\cite{chang2018rcaa} by 4.4\% and 4.5\% in mAP and top-1 accuracy, respectively. The proposed method also outperforms the single-task work, i.e. CNN$_{v}$+MGTS~\cite{Di2018Person}, in which each task is separately trained for person search. This demonstrates that joint training of multiple tasks is probable to outperform that of separated training of each task.

\begin{figure*}[t]
\begin{center}
   \includegraphics[height=10.5cm,width=19.0cm]{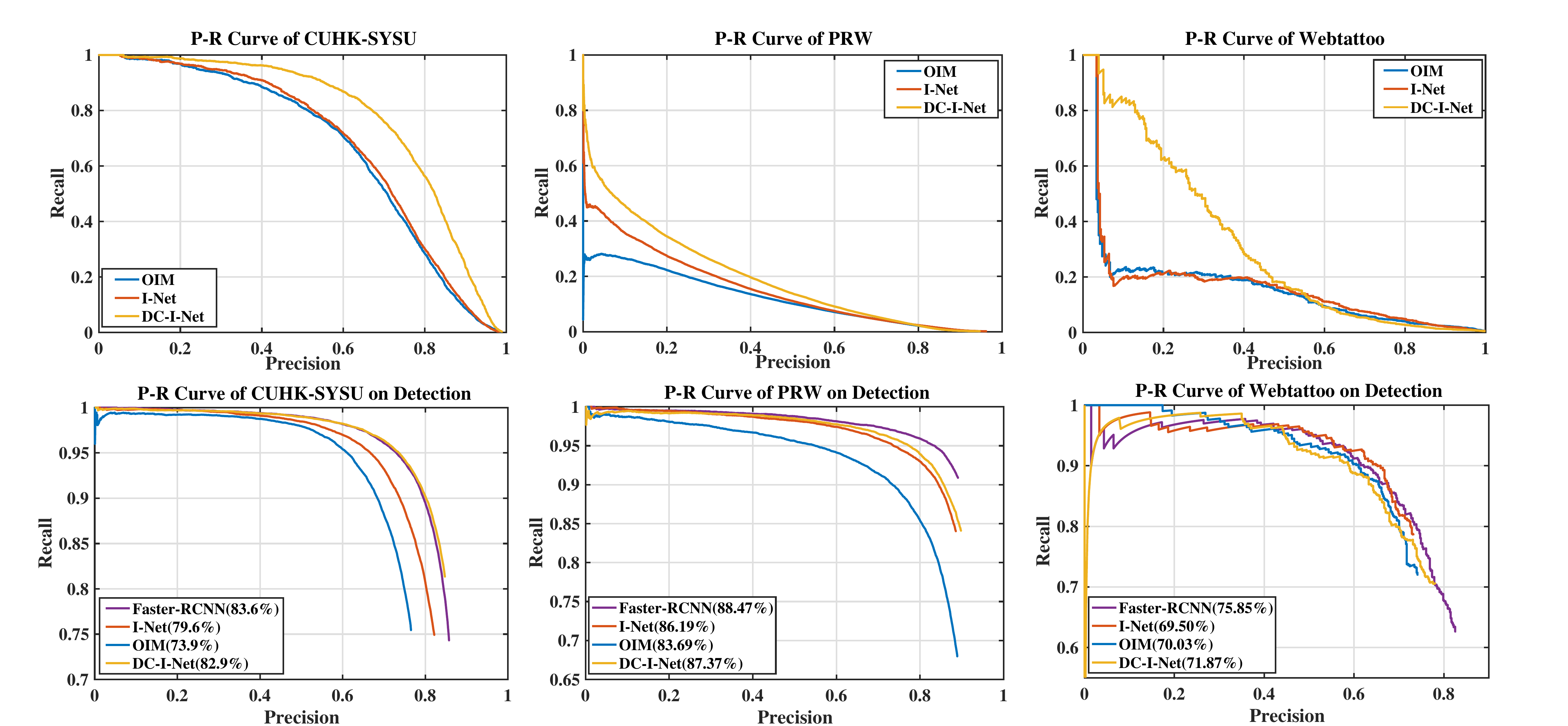}
\end{center}
   \caption{The Precision-Recall (P-R) curves of different methods on CUHK-SYSU (left), PRW (middle) and Webtattoo (right) datasets for object (i.e., person and tattoo) search and object detection, respectively. The first row shows the P-R curves of object search. The second row presents the P-R curves of detection, in which the AP value of detection is presented in the legend.}
\label{detection}
\end{figure*}

\begin{figure*}[t]
\begin{center}
   \includegraphics[width=1.0\linewidth]{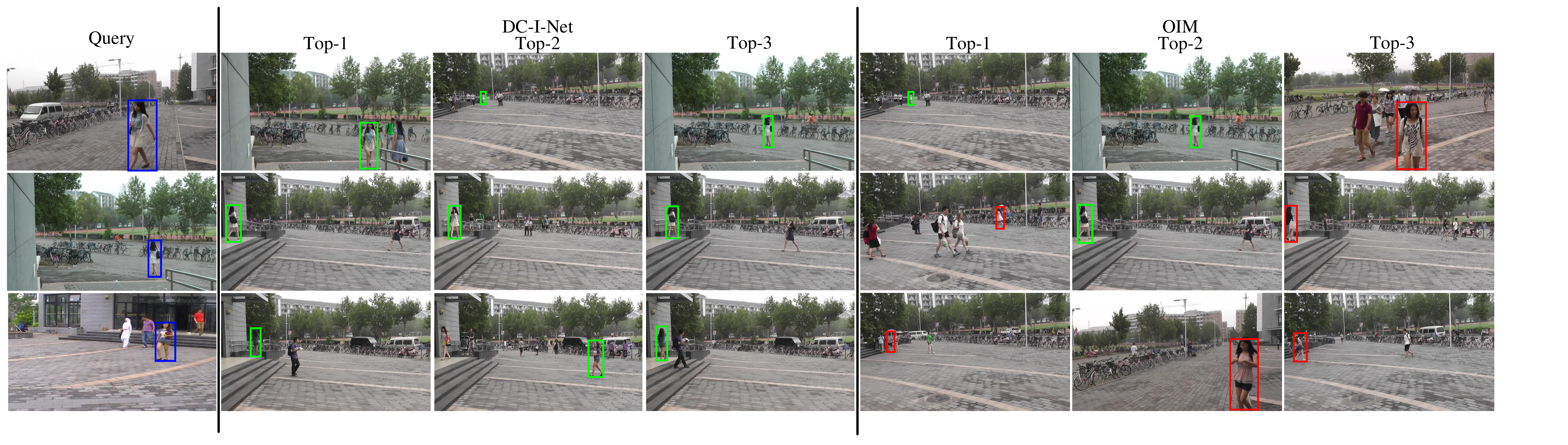}
\end{center}
   \caption{Retrieval of our DC-I-Net model (middle) and the OIM (right) with the PRW dataset by given three queries (left). The top 3 images with respect to the highest similarity scores of each model are shown. The blue boxes represent the target query person (probe), the green boxes mean correct matches, and the red boxes mean incorrect matches.}
\label{vis_prw}
\end{figure*}

\subsubsection{Visualization Results of Person Search}

Consider that OIM~\cite{Xiao_2017_CVPR} published in CVPR 2017 is the first work for person search and it also deployed a dictionary in the model, we therefore present the visualization results of the DC-I-Net and the OIM in Fig.~\ref{vis1} for comparison, in which eight queries are given and the person search results of top-1, top-2 and top-3 are shown. From the images, the difficulty of person search is observed because of the illuminations, resolutions and dense crowd. In the last two rows, the OIM fails to discover the target person in the top-3 results, while our DC-I-Net can still locate and match all the target persons correctly. From the visualization results, our model manifests better performance and robustness.

However, in some extreme conditions, our model also encounters false alarms (incorrect matches) as is shown in Fig.~\ref{vis2}. The extreme conditions are specified as follows. In Fig.~\ref{vis2} (a) and (d), the target persons stay in the crowds and overlapped with each other, which remains a difficult problem in person detection and Re-ID. In Fig.~\ref{vis2} (d) and (e), the person search is affected by illumination, an inherent factor, and the pose variations (b). The similar clothes in Fig.~\ref{vis2} (c), (e) and (f) can easily cause false alarms because of the inter-similarity. Since persons are not always clearly presented in real-world applications, the person search task still faces a challenge.

We further compare our DC-I-Net with the OIM and our previous I-Net on the CUHK-SYSU dataset from the performance of person search and pedestrian detection by presenting the Precision-Recall (P-R) curves in Fig.~\ref{detection} (left), respectively. The first row in Fig.~\ref{detection} denotes the P-R curves of person search performance and the second row represents the P-R curves of pedestrian detection performance. From the P-R curves, we see that the proposed DC-I-Net is much superior to other two closely-related state-of the art models for person search. Additionally, as shown in the second row of Fig.~\ref{detection} (left), the detection of the proposed DC-I-Net is comparable to the individually trained Faster-RCNN~\cite{Ren2017Faster}, \textcolor{black}{which represents the SOTA detection performance}. Therefore, the joint multi-task integration model is expected to have better performance beyond the single task of detection by handling multiple tasks appropriately.

\begin{table}[t]
\begin{center}
\caption{Comparisons of baselines, SOTA methods and our models on the PRW dataset}
\label{Table:PRW}
\begin{tabular}{|m{4cm}|m{1.5cm}|m{1.5cm}|}
\hline
Methods & mAP(\%) & Top-1(\%)  \\
\hline\hline
DPM~\cite{Felzenszwalb2010Object}+BOW~\cite{Zheng2015Scalable} & 9.7 & 31.1 \\
DPM~\cite{Felzenszwalb2010Object}+IDE$_{det}$~\cite{Zheng2016Person} & 18.8 & 45.9 \\
DPM-Alex+LOMO+XQDA~\cite{Liao2015Person} & 13.0 & 34.1  \\
DPM-Alex+IDE$_{det}$~\cite{Zheng2016Person} & 20.3 & 47.4  \\
DPM-Alex+IDE$_{det}$ + CWS~\cite{Zheng2016Person} & 20.5 & 48.3  \\
\hline
ACF~\cite{Dollar2014Fast}+LOMO+XQDA~\cite{Liao2015Person} & 10.5 & 30.9 \\
ACF~\cite{Dollar2014Fast}+IDE$_{det}$~\cite{Zheng2016Person} & 17.5 & 43.8 \\
ACF-Alex+LOMO+XQDA~\cite{Liao2015Person} & 10.3 & 30.6  \\
ACF-Alex+IDE$_{det}$~\cite{Zheng2016Person} & 17.5 & 43.6  \\
ACF-Alex+IDE$_{det}$ + CWS~\cite{Zheng2016Person} & 17.8 & 45.2  \\
\hline
LDCF~\cite{Nam2014Local}+BOW~\cite{Zheng2015Scalable} & 9.1 & 29.8 \\
LDCF~\cite{Nam2014Local}+LOMO+XQDA~\cite{Liao2015Person} & 11.0 & 31.1  \\
LDCF~\cite{Nam2014Local}+IDE$_{det}$~\cite{Zheng2016Person} & 18.3 & 44.6  \\
LDCF~\cite{Nam2014Local}+IDE$_{det}$+CWS~\cite{Zheng2016Person} & 18.3 & 45.5  \\
\hline
OIM~\cite{Xiao_2017_CVPR} & 21.3 & 49.9  \\
NPSM~\cite{Liu2017Neural} & 24.2 & 53.1  \\
I-Net & 25.6 & 48.7  \\
DC-I-Net(VGG16) & 30.4 & 53.3  \\
DC-I-Net(Resnet50) & \textbf{31.8} & \textbf{55.1}  \\
\hline
\end{tabular}
\end{center}
\end{table}

\subsection{Experiments on PRW Dataset}

\subsubsection{Compared Methods}
Similar to the CUHK-SYSU datasets, for the benchmark PRW dataset~\cite{Zheng2016Person}, we compare our DC-I-Net and I-Net with the SOTA methods for end-to-end person search such as OIM~\cite{Xiao_2017_CVPR} and NPSM~\cite{Liu2017Neural} and the baseline methods for person search with separate detection and re-id methods. Note that the DC-I-Net is trained with both VGG16 and Res50. Specifically, for separated detection, the DPM based~\cite{Felzenszwalb2010Object}, ACF based~\cite{Dollar2014Fast} and LDCF based~\cite{Nam2014Local} methods and their RCNN versions are considered. For separated Re-ID, the LOMO~\cite{Liao2015Person}+XQDA~\cite{Liao2015Person}, bag of words vector (BOW)~\cite{Zheng2015Scalable}, IDE$_{det}$, and CWS~\cite{Zheng2016Person} are considered. Therefore, 14 methods by combining the separated detection methods and separated Re-ID methods together are compared in this section. Note that in separated detection, for the RCNN based DPM and ACF detectors, AlexNet is implemented as the backbone according to~\cite{Zheng2016Person}.
\subsubsection{Experimental Results}
The results on the PRW dataset are shown in Table~\ref{Table:PRW}, from which we see that our DC-I-Net achieves the best results. \textcolor{black}{Specifically, our method outperforms the SOTA OIM~\cite{Xiao_2017_CVPR} by 10.5\% in mAP and 5.2\% in top-1 accuracy, which gets similar incremental with the experiments on the CUHK-SYSU~\cite{Xiao2016End}. Our method also outperforms the SOTA NPSM~\cite{Liu2017Neural} by 7.6\% and 2.0\% in mAP and top-1 accuracy, respectively}. It is worth noting that the end-to-end jointly trained models consistently outperform the separately trained models, which shows the effectiveness of joint multi-task learning for person search. Besides, our DC-I-Net is also much superior to I-Net, which demonstrates the effectiveness of the strategy of divide and conquer of each task in our joint multi-task integration framework.
The Precision-Recall (P-R) curves on PRW dataset in person detection and person search are presented in Fig.~\ref{detection} (middle). We observe that the proposed DC-I-Net shows the best performance for person search task. For detection task, \textcolor{black}{our model is also approaching the state-of-the-art performance of single task of Faster-RCNN detection~\cite{Ren2017Faster}}, and outperforms other SOTA models.

\subsubsection{Visualization Results of Person Search}
The person search results on PRW dataset are shown in Fig.~\ref{vis_prw}. The pose and resolution variations show the challenge of this dataset. From the top 3 retrieval images with respect to each query, we see that our proposed model can have much better search results than OIM. From the comparison, we see that our model is more robust to both pose and resolution variations.

\begin{figure}[t]
\begin{center}
   \includegraphics[width=1.0\linewidth]{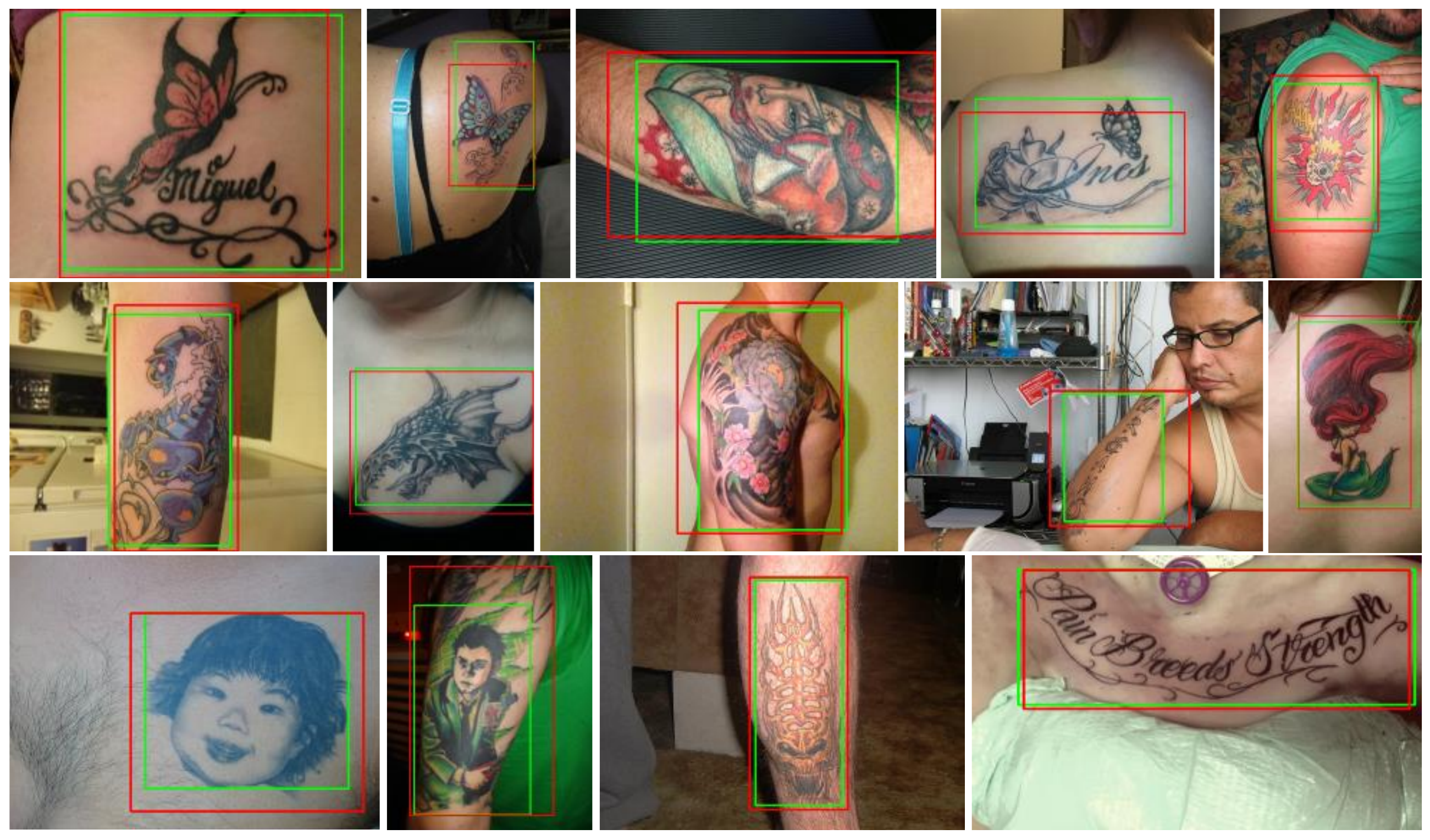}
\end{center}
   \caption{Detection results of the Webtattoo dataset by using DC-I-Net model. The red boxes represent the predicted bounding boxes and the green boxes represent the ground-truth bounding boxes.}
\label{web_detection}
\end{figure}

\subsection{Experiments on WebTattoo Dataset}

The Webtattoo dataset~\cite{han2019tattoo} is proposed search the images containing the same tattoos from the gallery image set as the query (probe) image. Therefore, this dataset is used for object (tattoo) search instead of person search. In this section, we present the results of tattoo detection and tattoo search. Considering that there is few work on tattoo search based on the compared models, for convenience, we compare with the SOTA OIM model~\cite{Xiao_2017_CVPR} because of its open-source advantage. In the experiments, the OIM~\cite{Xiao_2017_CVPR}, I-Net, and DC-I-Net are trained and tested without the 300K background tattoo images in the gallery set.
\subsubsection{Performance of Tattoo Detection}
The detection results based on our DC-I-Net are shown in Fig.~\ref{web_detection}, from which we could see that the detection of our model approaches the manually annotated ground-truth bounding boxes. Additionally, for comparison of the detection performance, the Precision-recall curves of different models on the Webtattoo dataset for detection are shown in Fig.~\ref{detection} (right, the second row). We see that the proposed DC-I-Net shows better performance than OIM and I-Net, and \textcolor{black}{it approaches the SOTA Faster-RCNN detection~\cite{Ren2017Faster}, a single task detection model}.

\subsubsection{Performance of Tattoo Search}
To show the importance of the detection part of end-to-end models, the results including mAP, top-1, top-5 and top-10 accuracies with and without (w/o) the detection part are presented in Table~\ref{Table:web}.

From this table, we observe that the proposed DC-I-Net significantly outperforms I-Net and OIM for both cases. Also, the models with the detection always get better performance than that without the detection part, because the detection enables the model focuses on the tattoo region in the images, such that the performances are dramatically increased sharply with the detection module. With the detection module, our DC-I-Net outperforms OIM and I-Net by 10.3\% and 6.8\% in mAP, and 11.5\% and 7.0\% in top-1 accuracy, respectively.
Further, the P-R curves of tattoo search on the test set as is shown in Fig.~\ref{detection} (right, the first row) also demonstrate the clear superiority of the DC-I-Net.

\begin{figure*}[t]
\begin{center}
   \includegraphics[width=0.9\linewidth]{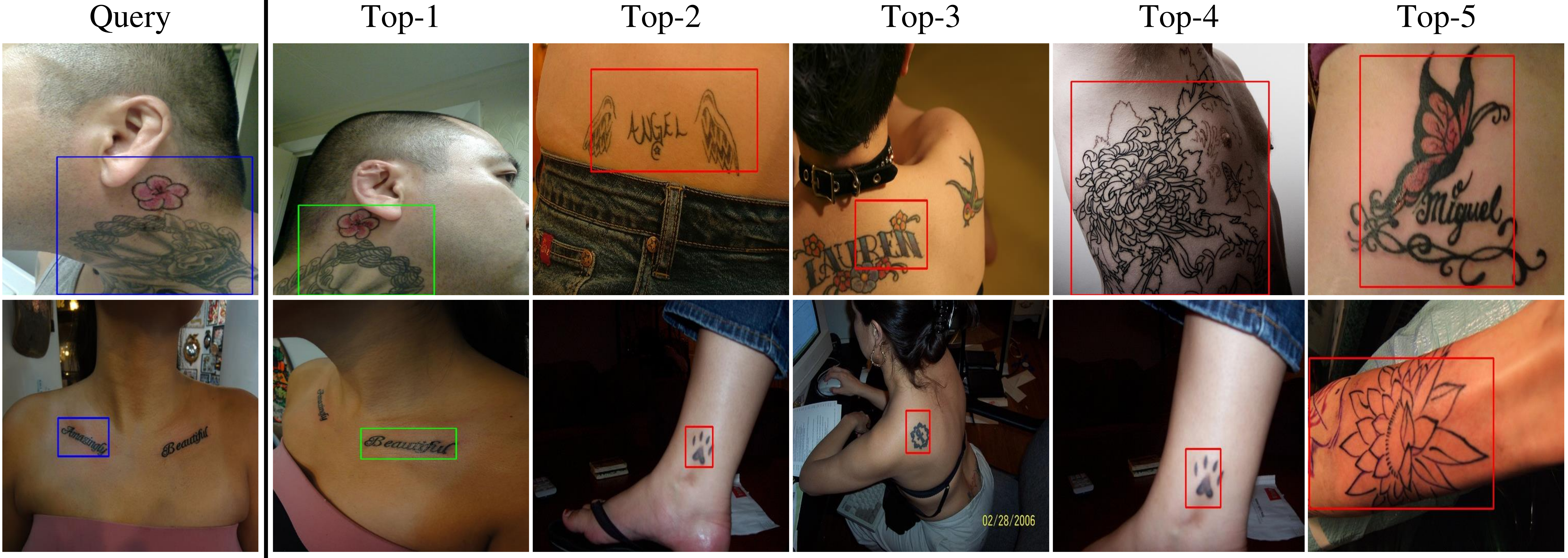}
\end{center}
   \caption{Retrieval results of our model with Webtattoo dataset. The top 5 retrieval results for each query are shown. The blue boxes represent the target query person (probe), the green boxes mean correct matches, and the red boxes mean incorrect matches.}
\label{vis_webtattoo}
\end{figure*}

The Tattoo image search results are visualized in in Fig.~\ref{vis_webtattoo}, in which the top 5 images with high similarity scores with respect to each query image are presented. The bounding boxes in the images are the automatically detected object regions of interest for matching and retrieval. Note that in image search task, the whole images rather than the object regions are fed into the model.

\begin{table}[t]
\begin{center}
\caption{Tattoo search results of different models on Webtattoo dataset}
\label{Table:web}
\begin{tabular}{|m{2cm}|c|c|c|c|}
\hline
With Detection& mAP(\%) & Top-1(\%) & Top-5(\%) & Top-10(\%) \\
\hline\hline
OIM~\cite{Xiao_2017_CVPR} & 38.2 & 39.5 & 55.5 & 60.5 \\
\hline
I-Net & 41.7 & 44.0 & 61.0 & 67.5 \\
\hline
DC-I-Net & \textbf{48.5} & \textbf{51.0} & \textbf{66.5} & \textbf{71.5} \\
\hline
\hline
w/o Detection& mAP(\%) & Top-1(\%) & Top-5(\%) & Top-10(\%) \\
\hline
OIM~\cite{Xiao_2017_CVPR} & 21.5 & 23.0 & 40.0 & 41.5 \\
\hline
I-Net & 23.5 & 25.5 & 39.5 & 44.0 \\
\hline
DC-I-Net & \textbf{30.4} & \textbf{31.0} & \textbf{51.5} & \textbf{64.0} \\
\hline
\end{tabular}
\end{center}
\end{table}

\section{Model Analysis and Discussion}
\label{modelanalysis}
In the section, to have a deep insight on the effectiveness of the proposed models, the model analysis and discussion are presented based on the CUHK-SYSU~\cite{Xiao2016End} dataset.

\subsection{Ablation Study of Model Losses}
\subsubsection{Discussion on the HEP and C$^2$HEP Losses}
Compared to the I-Net, two key improvements are made in DC-I-Net, including the new network structure and the identity classification loss (i.e., C$^2$HEP). We therefore conduct study of the performance gap between HEP in I-Net and C$^2$HEP in DC-I-Net. By alternatively changing the softmax guided identity classification loss in each network, the results are presented in Table~\ref{Table:ablation}. From the table, we observe that by changing the HEP loss into the C$^2$HEP loss in the I-Net, the mAP and top-1 accuracy are improved by 1.4\% and 1.9\%, respectively. Similarly, by changing the C$^2$HEP loss into HEP loss in the DC-I-Net, the mAP and top-1 accuracy are degraded by 2.7\% and 2.8\%, respectively. These results demonstrate the effectiveness of the newly proposed C$^2$HEP loss over HEP loss. Note that the HEP is in default in I-Net and C$^2$HEP is in default in DC-I-Net. By comparing the results between I-Net (C$^2$HEP) and DC-I-Net, or between I-Net and DC-I-Net (HEP), the increased performance clearly reflects the advantage benefiting from the newly proposed architecture in DC-I-Net.

\begin{table}[t]
\begin{center}
\caption{Performance comparison between I-Net and DC-I-Net based on HEP and C$^2$HEP losses, respectively. Note that without special indication, HEP is deployed in I-Net and C$^2$HEP is deployed in DC-I-Net.}
\label{Table:ablation}
\begin{tabular}{|m{2.1cm}|c|c|c|c|}
\hline
Loss Type& mAP(\%) & Top-1(\%) & Top-5(\%) & Top-10(\%) \\
\hline\hline
I-Net & 79.5 & 81.5 & 92.2 & 94.6 \\
\hline
I-Net (C$^2$HEP) & 80.9 & 83.4 & 94.1 & 95.2 \\
\hline
DC-I-Net (HEP) & 81.0 & 83.0 & 93.2 & 95.6 \\
\hline
DC-I-Net & \textbf{83.7} & \textbf{85.8} & \textbf{94.3} & \textbf{96.1} \\
\hline
\end{tabular}
\end{center}
\end{table}

\begin{figure*}[t]
\begin{center}
   \includegraphics[height=5.0cm,width=18.5cm]{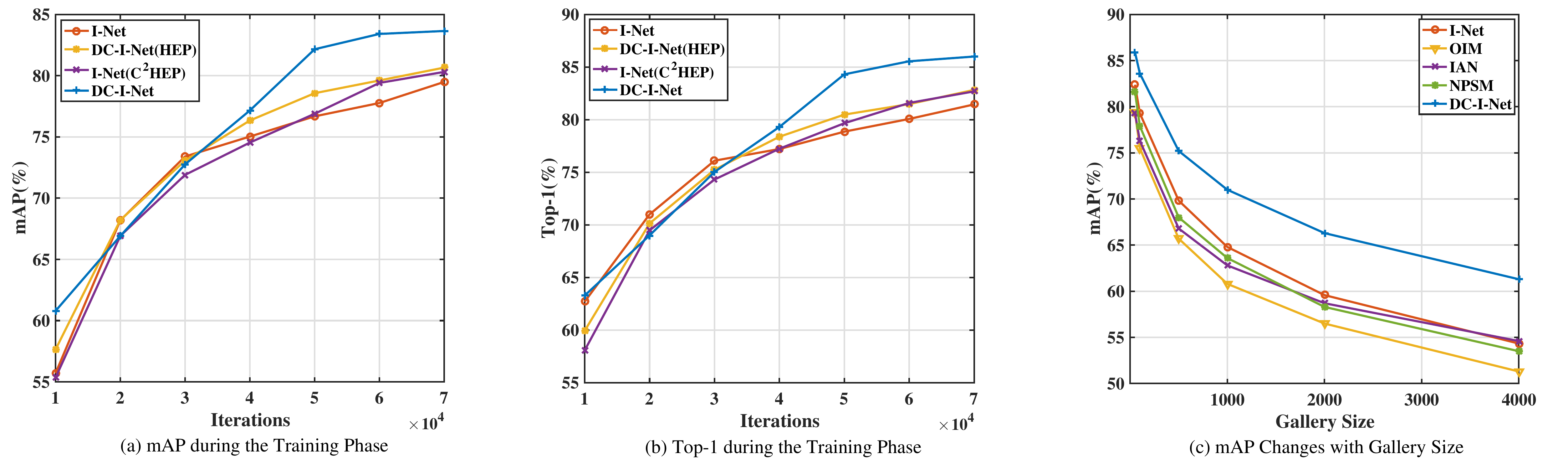}
\end{center}
   \caption{Discussions on the model. (a) Performance variation (mAP) with different gallery sizes for different models. (b) Performance comparison (mAP) between DC-I-Net and I-Net under different identity classification losses (C$^2$HEP and HEP) with respect to each iteration during the training phase. (c) Performance comparison (top-1 accuracy) between DC-I-Net and I-Net under different identity classification losses (C$^2$HEP and HEP) with respect to each iteration during the training phase. Note that without special indication, I-Net is deployed with HEP loss and DC-I-Net is deployed with C$^2$HEP loss in default.}
\label{figs}
\end{figure*}

\textit{Remarks}. \textcolor{black}{We analyze here the reason why C$^2$HEP is effective. The HEP and C$^2$HEP losses are supervised by the ground-truth identities.} Since person search is an open set problem, we observe the the identity classification accuracy of the training set during iterations. The HEP and the traditional softmax loss are trained for comparing with our C$^2$HEP loss. For fair comparison, the ground-truth bounding boxes of only labeled target persons in each image are cropped and fed into the model to get the classification accuracy in each iteration. The performance variations based on HEP and C$^2$HEP with iterations during the training phase are shown in Fig.~\ref{figs} (a) and (b), from which we see that the newly proposed C$^2$HEP loss significantly outperforms the HEP during the training phase, for both network architectures. We know that in the CUHK-SYSU dataset, 5532 different identities are labeled for training. In each iteration, \textcolor{black}{the input images only contain a very few identities (few-shots) due to the small number of input images}, which means that the model training should require thousands of iterations to traverse almost all the identities of the dataset. Therefore, the weights of the old HEP loss is hard to be properly trained, which thus leads to worse performance. In the new C$^2$HEP loss, by taken into account the class centers computed via the input features, the identity discrimination can be easily captured because of full-shot property.

\begin{table}[t]
\begin{center}
\caption{\color{black}{Performance comparisons of different joint losses: metric loss and identity discrimination loss based on the DC-I-Net architecture}}
\label{Table:HEPsoft}
\begin{tabular}{|m{2.0cm}|c|c|c|c|}
\hline
Loss Type& mAP(\%) & Top-1(\%) & Top-5(\%) & Top-10(\%) \\
\hline
\hline
Triplet+HEP & 67.8 & 69.6 & 87.6 & 92.2 \\
\hline
OLP only & 81.3 & 82.9 & 93.9 & 96.0 \\
\hline
C$^2$HEP only & 82.2 & 84.7 & 94.0 & 96.0 \\
\hline
OLP + HEP & 81.9 & 83.9 & 93.9 & 95.6 \\
\hline
OLP + C$^2$HEP & \textbf{83.7} & \textbf{85.8} & \textbf{94.3} & \textbf{96.1} \\
\hline
\end{tabular}
\end{center}
\end{table}

\begin{figure}[h]
\begin{center}
   \includegraphics[width=1.0\linewidth]{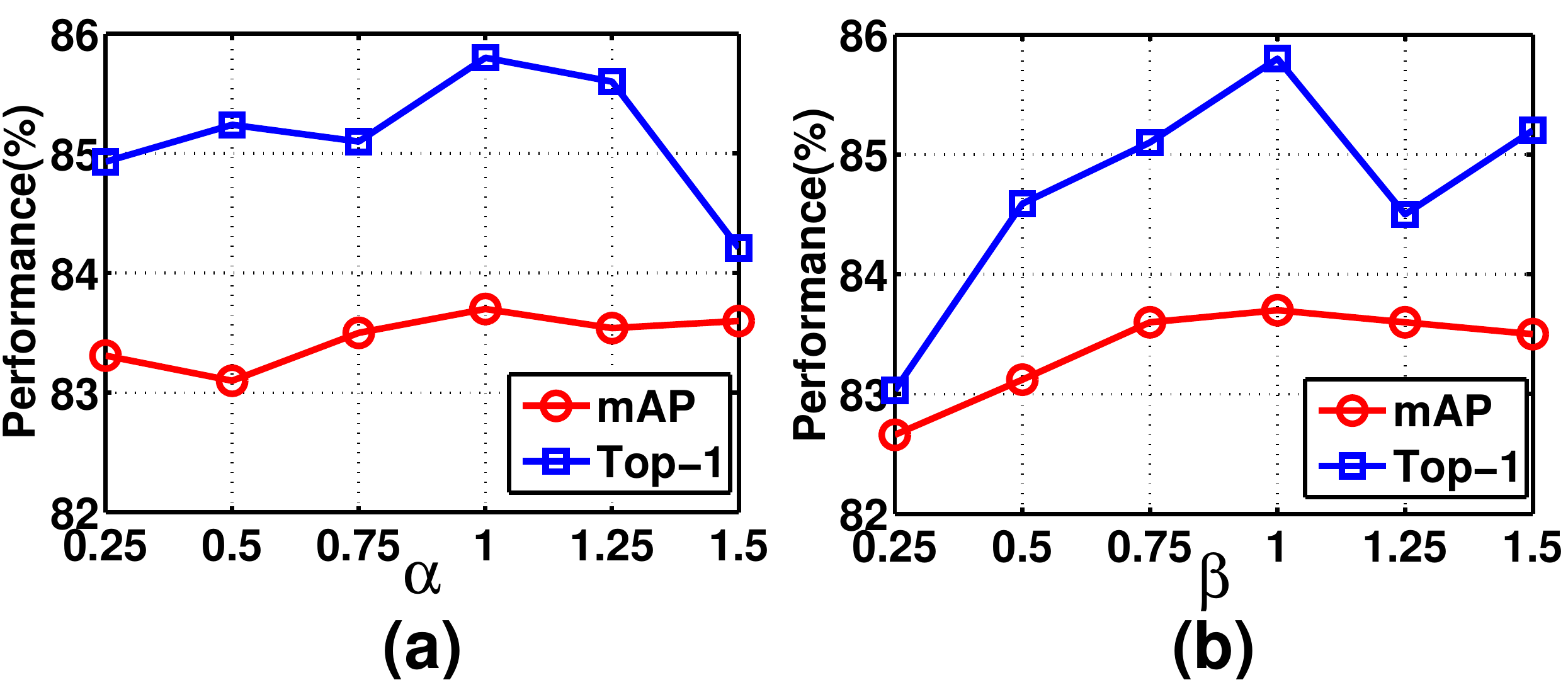}
\end{center}
   \caption{\textcolor{black}{Sensitivity analysis of the loss weights in Eq.~(\ref{TotalLoss}). We fix one parameter and adjust another one in experiments. (a) $\beta=1$; (b) $\alpha=1$.}}
\label{loss_weight}
\end{figure}

\subsubsection{Discussion on Joint Metric Loss and Identity Loss}
In order to study the effectiveness of the proposed losses, including the metric based OLP loss and the identity based HEP/C$^2$HEP loss. For comparison, we train our network based on different combinations of the metric loss (triplet loss, OLP) and identity loss (HEP, C$^2$HEP). Note that the network structure is the same as the DC-I-Net. The results are presented in Table~\ref{Table:HEPsoft}.

From the table, we see that the traditional triplet loss together with the softmax guided cross-entropy loss achieves much worse performance in mAP and top-1 accuracy than ours. With the proposed OLP loss only or C$^2$HEP loss only, much better performances can be achieved. The worse performance of triplet loss is mainly caused by the stagnation problem during training the end-to-end person search model. Therefore, the proposed softmax guided OLP loss can well overcome the stagnation problem of the triplet loss. With the proposed C$^2$HEP based identity loss only, the performance is comparable to OLP based metric loss only. By combining them together, the proposed DC-I-Net achieves the best performance, 83.7\% and 85.8\% in mAP and top-1 accuracy, respectively. The effectiveness of the proposed losses are verified in this section.

\subsubsection{The Analysis of Loss Weight}
\textcolor{black}{In this section, we adjust the weights of OLP loss and C$^2$HEP loss of DC-I-Net during the training phase. We fix the $\alpha$ as 1.0 in the Eq.~(\ref{TotalLoss}) and adjust the $\beta$ from 0.25 to 1.5 to study the influence and vice versa. The backbone of DC-I-Net is set as VGG16 in the experiments. The mAP and Top-1 with respect to different loss weights are shown in the Fig.~\ref{loss_weight}. From the results, we observe that the trade-off parameters have slight impact and the best performance can be easily achieved when $\alpha=1$ and $\beta=1$.}

\begin{table}[t]
\begin{center}
\caption{Performance with different feature dictionary size stored in OLP based on the DC-I-Net architecture.}
\label{Table:bg}
\begin{tabular}{|c|m{2cm}|m{1.5cm}|m{1.5cm}|}
\hline
mAP (\%) & $20\times128$ & $40\times128$ & $60\times128$ \\
\hline
OLP only & 81.4 & 81.3 & 81.8 \\
\hline
OLP+C$^2$HEP & 82.5 & \textbf{83.7} & 83.1 \\
\hline\hline
Top-1 (\%) & $20\times128$ & $40\times128$ & $60\times128$ \\
\hline
OLP only & 83.5 & 82.9 & 83.4 \\
\hline
OLP+C$^2$HEP & 84.7 & \textbf{85.8} & 85.2 \\
\hline
\end{tabular}
\end{center}
\end{table}

\subsection{Analysis of the Feature Dictionary in OLP and Priority Classes in C$^2$HEP}
\subsubsection{Impact of the Feature Dictionary}
In the proposed OLP loss, a feature dictionary is deployed to generate negative pairs, which may preferably restrict the positive pairs for effective metric learning. Therefore, the impact of the dictionary size, i.e. the number of features stored, is studied. The model is trained either with only OLP or with both OLP and C$^2$HEP loss.

The number of features stored in the feature dictionary of OLP depends on the minibatch size (i.e., 128). In this study, different times (i.e., 20, 40 and 60) the minibatch size are determined as the dictionary size (i.e., 20$\times$128, 40$\times$128, 60$\times$128) and tested, respectively. Note that 40$\times128$ means that the number of features stored in the dictionary is 40 times the minibatch size (128), i.e. 5120. The analysis results are presented in Table~\ref{Table:bg}, in which we see that the impact of the dictionary size is slight. With both the OLP and C$^2$HEP loss functions implemented, i.e., DC-I-Net, the model gets the best results when the dictionary size is set as 40$\times128$.

From Table~\ref{Table:bg}, two perspectives can be observed.
1) From the case of OLP loss only trained model, increasing the feature dictionary size does not improve the performance, which is because the OLP loss is hard to contain all identities in the dataset and even the number of stored features grows, the number of identities in the feature dictionary is not significantly increased.
2) From the case of both OLP and C$^2$HEP losses trained model, the best performance with a suitable size of feature dictionary can be achieved, which is because the C$^2$HEP loss can explore all the labeled identities in the dataset. As a result, the dictionary size is set as 40$\times128$ in experiments.

\subsubsection{Impact of the Number of Priority Classes}

The essence of C$^2$HEP lies in the class center guided hard example priority mechanism. Therefore, utilizing different number of priority classes for computing the C$^2$HEP loss is studied in experiments. Note that, in this study, both OLP and C$^2$HEP loss functions are used for implementation. Specifically, in the experiments, the number of the selected priority classes is set as 50, 100, 1000, and 5532, respectively. Note that the number 5532 means that all the labeled persons are used in loss computation without consideration of the priority classes. The results are presented in Table~\ref{Num_pri}, from which we observe that the model trained with 100 selected priority classes outperforms the model without the hard example priority strategy (i.e., 5532) by 0.8\% in mAP and 2.1\% in top-1 accuracy. The effectiveness of the proposed hard example priority strategy is verified. Additionally, with the increasing of the number of priority classes, the performance is slightly degraded which is due to that the attention on the really hard classes may be reduced. In contrast, if the number of priority classes is too small, the model may not well explore the identities of the whole dataset such that the performance is not good. Therefore, a suitable number of priority classes is required, which is set as 100 in experiments.

\begin{table}[t]
\begin{center}
\caption{Performance comparisons by using different numbers of selected priority classes based on the DC-I-Net architecture.}
\label{Num_pri}
\begin{tabular}{|m{1.5cm}|c|c|c|c|}
\hline
~ & mAP(\%) & Top-1(\%) & Top-5(\%) & Top-10(\%) \\
\hline\hline
50 & 82.5 & 84.8 & 94.1 & 96.0 \\
\hline
100 & \textbf{83.7} & \textbf{85.8} & 94.3 & \textbf{96.1} \\
\hline
1000 & 83.1 & 85.0 & \textbf{94.6} & 95.8 \\
\hline
5532 & 82.9 & 83.7 & 93.8 & 95.6 \\
\hline
\end{tabular}
\end{center}
\end{table}

\subsection{Analysis of the Retrieval Performance}
\subsubsection{Impact of Gallery Size}

The person search is essentially an image retrieval task, which therefore becomes more challenging especially when the size of gallery set (retrieval pool) increases. This section presents a study on the impact of different gallery size. Specifically, in the experiments, we vary the gallery size as 50, 100, 500, 1000, 2000, and 4000, respectively, and test the mAP of different models including DC-I-Net, I-Net, the OIM~\cite{Xiao_2017_CVPR}, NPSM~\cite{Liu2017Neural}, and IAN~\cite{Xiao2017IAN} for each gallery size. The retrieval performance variation curves are shown in Fig.~\ref{figs} (c), from which we can see that with the increasing gallery size of the test set, the performances of all methods are degraded. It is worthy noting that our proposed DC-I-Net has a relatively slower degradation speed, but always outperforms other methods with respect to each gallery size.
\begin{table}[h]
\begin{center}
\caption{\textcolor{black}{Performance analysis with different numbers of input images at every iteration based on different losses.}}
\label{mul_input}
\begin{tabular}{|c|c|c|c|c|}
\hline
~ & \multicolumn{2}{|c|}{OLP+C$^2$HEP} & \multicolumn{2}{|c|}{Contrastive Loss} \\
\cline{2-5}
~ & mAP(\%) & Top-1(\%) & mAP(\%) & Top-1(\%) \\
\hline\hline
2-input & \textbf{86.2} & \textbf{86.5} & 48.7 & 45.0 \\
\hline
4-input & \textbf{85.9} & \textbf{87.0} & 54.4 & 54.7 \\
\hline
8-input & \textbf{85.8} & \textbf{85.9} & 60.4 & 60.7 \\
\hline
\end{tabular}
\end{center}
\end{table}
In fact, it is common that the difficulty in finding the query person is growing with the increasing gallery size because of the enhanced inter-similarity.

\subsubsection{Performance Variation With Training Iterations}
This section presents the retrieval performance variation of mAP and top-1 accuracy by using the proposed DC-I-Net and the previous I-Net, both of which are trained for 70K iterations. The retrieval performance variation of our models during the training phase is shown in Fig.~\ref{figs} (a) and (b), where both the mAP and top-1 accuracy increase with the training iterations. DC-I-Net shows faster upward trend than I-Net. Note that the learning rate is reduced at the 40K iteration in the training phase.

\subsection{Analysis of Different Numbers of Input Images}
\textcolor{black}{The insufficient number of input images for each iteration leads to the stagnation problem for the traditional loss function (e.g. contrastive loss). In this part, we change the number of input images of each iteration for ablation analysis. The resnet50 based DC-I-Net architecture is implemented based on the proposed losses and the contrastive loss. Specifically, we set the number of input training images as 2, 4, and 8, respectively in each iteration in this experiment. The results are shown in Table~\ref{mul_input}, from which we can clearly see that when the number of input images is small, the traditional contrastive loss even does not work. The reason is that the model based on contrastive loss encounters the training stagnation problem when the number of input images is set as 2. As the number of input images increases, the stagnation problem is alleviated and the performance is progressively improved.}

\textcolor{black}{On the contrary, our proposed losses always work well. The proposed OLP and C$^{2}$HEP loss functions have well reduced the influence of the stagnation problem, so that the number of input images does not have much impact on our model. This fully verifies the motivation and effectiveness of the proposed OLP and C$^2$HEP losses.}

\section{Conclusion}
\label{conclusion}
In this paper, we propose an end-to-end tasks-integrated network (I-Net) for user-friendly image search, by jointly modeling the detection and retrieval tasks in a unified framework. While I-Net does not consider the task specification and the \textcolor{black}{inherent problem of small number of input images in training}, we have further proposed an essentially improved integrated network with the philosophy of divide and conquer, called DC-I-Net. Two merits are naturally formulated: 1) the task specification can be explored and more accurate object proposals are used to effectively train the re-identifier by deploying the detector in front of the re-identifier. 2) A novel class center guided hard example priority (C$^2$HEP) loss is proposed by utilizing the class centers computed via the timely updated input features, which well overcomes the inherent few-shot (i.e., very few identities) problem during training of image search model. The proposed models outperform the state-of-the-art task-integrated and task-separated image search models on three widely used benchmark datasets, such as CUHK-SYSU~\cite{Xiao2016End}, PRW~\cite{Zheng2016Person} and Webtattoo ~\cite{han2019tattoo}.

\ifCLASSOPTIONcaptionsoff
  \newpage
\fi

%


%
%

\ifCLASSOPTIONcaptionsoff
  \newpage
\fi



\bibliographystyle{IEEEtran}
\bibliography{egbib1}
\end{document}